\documentclass[11pt]{article}

\usepackage[preprint]{acl}

\usepackage{times}
\usepackage{latexsym}
\usepackage[T1]{fontenc}
\usepackage[utf8]{inputenc}
\usepackage{microtype}
\usepackage{inconsolata}

\usepackage{verbatim}
\usepackage{hyperref}
\usepackage{soul}
\usepackage{graphicx}
\usepackage{amsmath}
\usepackage{amssymb}
\usepackage{amsthm}
\usepackage{mathtools}
\usepackage{array}
\usepackage{booktabs}
\usepackage{multirow}
\usepackage{bbm}
\usepackage{enumitem}
\usepackage{todonotes}
\usepackage{tabularx}
\usepackage{xcolor}

\newcolumntype{x}[1]{>{\centering\let\newline\\\arraybackslash\hspace{0pt}}p{#1}}

\newcommand{\SectionSpacing}{0pt}
\newcommand{\SubsectionSpacing}{0pt}
\newcommand{\ParagraphHeaderSpacing}{\par\medskip\noindent}

\newcolumntype{L}[1]{>{\raggedright\let\newline\\\arraybackslash\hspace{0pt}}m{#1}}
\newcolumntype{C}[1]{>{\centering\let\newline\\\arraybackslash\hspace{0pt}}m{#1}}
\newcolumntype{R}[1]{>{\raggedleft\let\newline\\\arraybackslash\hspace{0pt}}m{#1}}

\theoremstyle{definition}

\newcommand{\name}{BIRD-Persona}
\newcommand{\base}{BIRD-Python}
\newcommand{\tts}{Text-to-SQL}
\newcommand{\ttp}{Text-to-Python}

\usepackage[acronym]{glossaries}
\makeglossaries

\glsdisablehyper
\newacronym{ps}{PS}{Persona Score}
\newacronym{scs}{SCS}{Structural Complexity Score}
\newacronym{llm}{LLM}{large language model}

\makeatletter
\AtBeginDocument{%
  \patchcmd{\@maketitle}
    {\vskip 0.2in plus 1fil minus 0.1in}
    {\vskip 0.28in plus 1fil minus 0.1in}
    {}{}%
  \patchcmd{\@maketitle}
    {\vskip 0.3in plus 2fil minus 0.1in}
    {\vskip 0.45in plus 2fil minus 0.1in}
    {}{}%
}
\makeatother

\title{\name{}: A Persona-Aware Benchmark for \\SQL and pandas Analytics Code Generation}

\author{Davin Hill\textsuperscript{*} \\
  Optum AI \\
  \texttt{davin\_hill@optum.com} \\\And
  Pawan Chorasiya\textsuperscript{*} \\
  Optum AI \\
  \texttt{pawan\_chorasiya@optum.com} \\\And
  Bhawani S. Leelar \\
  Optum AI \\
  \texttt{bhawani.leelar@optum.com} \\\AND
  Oren Avram \\
  Optum AI \\
  \texttt{oren.avram@optum.com} \\\And
  Robert E. Tillman \\
  Optum AI \\
  \texttt{rob.tillman@optum.com} \\\And
  Tamer Soliman \\
  Optum AI \\
  \texttt{tamer.soliman@optum.com} \\}

\begin{document}
\maketitle

\begingroup
\renewcommand{\thefootnote}{*}
\footnotetext{Equal contribution.}
\endgroup
\setcounter{footnote}{0}

\begin{abstract}
Production text-to-code systems increasingly use \glspl{llm} to translate natural-language analytics requests into executable SQL and pandas code, serving both business users, who describe desired results in application terms, and technical users, who name tables and columns directly.
Existing \tts{} and \ttp{} benchmarks provide coverage, but they do not isolate when SQL and pandas performance diverge.
We introduce \name{}, a persona-stratified extension of \base{} that adds novice/expert rewrites (business-style and technical-style phrasing) to tasks with matched SQL and pandas ground truth, plus schema variants and task-structure annotations.
Across 1,534 tasks, nine models, and 245,592 evaluations, \name{} reproduces the previously reported SQL-over-Python gap: SQL outperforms pandas for eight of nine systems.
The gap is not uniform: expert, schema-explicit prompts improve both paradigms but help SQL more, schema representations have mixed model-dependent effects, and selected-column burden widens the gap while subqueries do not. By making user phrasing a controlled benchmark dimension, \name{} provides a foundation for persona-aware and paradigm-aware evaluation of natural-language analytics systems. The \name{} dataset is available on \href{https://huggingface.co/datasets/optum-cognition-lab/BIRD-Persona}{Huggingface}.
\end{abstract}

\vspace{\SectionSpacing}
\section{Introduction}
\vspace{\SectionSpacing}

\glspl{llm} are increasingly used to translate natural-language data requests into executable code, letting users retrieve and analyze data without writing programs by hand. In user-facing analytics systems, the same database may serve both business users, who describe desired outcomes in application terms, and technical users, who name tables and columns directly. The generated code typically follows one of two paradigms: \tts{}, which targets relational databases, and \ttp{}, where the Python target is pandas code over tabular data. Because SQL expresses relational operations declaratively while pandas requires step-by-step dataframe operations, a matched benchmark that pairs SQL queries and pandas code for the same underlying task can measure when user phrasing and target paradigm affect reliability.

Existing benchmarks provide important coverage for each paradigm, but they do not show which user, system, and task conditions widen or narrow the SQL-pandas gap. \tts{} benchmarks such as Spider \citep{yu2018spider}, BIRD \citep{li2023bird}, and Spider 2.0 \citep{lei2025spider2} cover increasingly realistic database-grounded and enterprise settings, while DSCodeBench \citep{ouyang2026dscodebench} evaluates realistic Python data-science code generation. Closest to our setting is \base{}  \citep{huBenchmarkingTexttoPythonTexttoSQL2026}, which aligns each question with both SQL and pandas solutions, allowing \tts{} and \ttp{} to be compared on the same analytics task. The user condition is especially relevant: prior work shows that user expertise affects satisfaction and engagement in general \gls{llm} interactions \citep{palta-etal-2025-speaking}, further motivating a persona-aware evaluation axis for code generation. A more detailed summary of related works is provided in App. \ref{app:background:related_works}.

Motivated by these gaps, we introduce \name{}, a persona-stratified extension of \base{}. Starting from 1,534 \base{} tasks with matched SQL and pandas ground truth, we construct novice/expert rewrites of each question, corresponding to business-style and technical-style phrasing, while preserving the intended answer. 
We further add schema variants and task-structure annotations, including structural-complexity features and operation tags, to support diagnostic evaluation beyond aggregate accuracy. We evaluate nine \glspl{llm} across SQL and pandas targets, original, business-style, and technical-style persona conditions, and three schema representations.

We use \name{} to show how the SQL-pandas gap changes across user, task, and system conditions. After reporting descriptive accuracy by model, user phrasing, schema representation, and task structure, we fit mixed-effects logistic models to estimate whether phrasing specificity, structural complexity, and schema representation are associated with that gap. This moves beyond a single leaderboard score toward a practical question: when is one paradigm more reliable than the other?

Using \name{}, we first reproduce the previously reported SQL-over-Python gap, validating the benchmark against prior cross-paradigm results, and then reveal three patterns that aggregate accuracy obscures. First, technical-style, schema-explicit phrasing improves both paradigms, but benefits SQL more than pandas. Second, richer schema representations produce mixed, model-dependent effects rather than a consistent improvement. Third, aggregate structural complexity captures broad difficulty but misses finer effects; selected-column burden and subquery structure shift the SQL-pandas gap in opposite directions.
In summary, we have the following \textbf{main contributions:}
\begin{itemize}[leftmargin=*, itemsep=-1pt]
    \item \textbf{\name{}}, to our knowledge the first benchmark to make persona-based user phrasing a controlled dimension for matched SQL and pandas analytics code generation, augmented with schema variants, structural-complexity features, and operation tags.
    \item An evaluation of nine \glspl{llm} across SQL and pandas targets, persona conditions, and schema representations, reproducing the previously reported SQL-over-Python gap.
    \item A diagnostic analysis showing that the SQL-pandas gap varies with phrasing specificity, schema representation, and task structure, revealing patterns hidden by aggregate accuracy alone.
\end{itemize}

\vspace{\SectionSpacing}
\section{Methods}
\vspace{\SectionSpacing}
\label{sec:methods}

We build on \base{}, which pairs each natural-language analytics question with semantically aligned SQL and pandas implementations \citep{huBenchmarkingTexttoPythonTexttoSQL2026}. Using this corpus, we construct a benchmark and evaluation protocol for comparing target paradigms under controlled variation in user phrasing and schema representation, with task-structure annotations for diagnostic analysis. We then use this benchmark to analyze when performance differs between SQL and pandas.

\vspace{\SubsectionSpacing}
\subsection{Dataset Construction}
\vspace{\SubsectionSpacing}
\label{subsec:dataset-construction}

Our base corpus contains 1,534 \base{} source tasks. Each item includes an original natural-language question, database identifier, evidence string, gold SQL, gold pandas code, and BIRD difficulty label. The task, not the individual program, is the unit of analysis: the SQL and pandas programs are two implementations of the same intended request. We extend this corpus with user-phrasing variants, schema-representation variants, and task-structure annotations, described next.

\subsubsection{Persona Paraphrase Generation}

\begin{table}[t]
\centering
\scriptsize
\setlength{\tabcolsep}{3pt}
\renewcommand{\arraystretch}{1.08}
\begin{tabularx}{\columnwidth}{@{}l>{\raggedright\arraybackslash}X>{\raggedright\arraybackslash}X@{}}
\toprule
\textbf{Dimension} & \textbf{Novice} & \textbf{Expert} \\
\midrule
Schema/Scope & Business terms & Table and column names \\
\cmidrule(lr){1-3}
Values & Approximate dates or values & Formats, units, constants \\
\cmidrule(lr){1-3}
Logic & Implicit grain, ties, nulls & Explicit grain, ties, nulls \\
\cmidrule(lr){1-3}
\shortstack[l]{Output\\Specificity} & Minimal presentation detail & Sort, limits, column order \\
\bottomrule
\end{tabularx}
\vspace{-2mm}
\caption{Persona rewrite controls used to separate business-style and schema-explicit phrasings while preserving the intended answer.}
\label{tab:persona-rewrite-controls}
\vspace{-2mm}
\end{table}

To vary user phrasing, we give each source task a novice rewrite and an expert rewrite, corresponding to business-style and technical-style phrasing, both intended to preserve the original answer. Together with the unmodified \base{} question, these form three persona conditions: original, novice, and expert. The rewrites differ along four controlled specificity dimensions, shown in Table \ref{tab:persona-rewrite-controls}. We use novice and expert as shorthand for these two phrasing registers, business-style/underspecified versus schema-explicit, and they describe how a request is phrased rather than the cognitive expertise or proficiency of the person issuing it.

To quantify phrasing specificity, we score each question with the \gls{ps}, a lexical-grounding measure of how explicitly the question mentions schema, value, and operator tokens from the gold SQL. For each gold SQL query, we extract a token set \(T\) containing table names, column names, string literals, dates, and numeric constants, then normalize these tokens by lowercasing, removing quotes, and replacing identifier separators such as underscores with spaces. For a question \(q\), we construct candidate \(n\)-grams \(C(q)\) with \(n \in \{1,2,3\}\) and compute character-level Ratcliff-Obershelp similarity \citep{ratcliff1988pattern} between each SQL token \(t \in T\) and each candidate phrase \(c \in C(q)\). Greedy one-to-one matching then assigns each SQL token a real-valued score \(r_t \in [0,1]\), using each question token at most once. Here \(r_t=1\) indicates an exact lexical match, partial overlaps receive fractional credit, and unmatched SQL tokens receive \(r_t=0\). 

\ParagraphHeaderSpacing
The \gls{ps} is:
\setlength{\abovedisplayskip}{2pt}
\setlength{\belowdisplayskip}{2pt}
\[
  \mathrm{PS}(q,\mathrm{SQL})
  = \frac{1}{|T|}\sum_{t\in T} r_t^{\alpha}
\]
where \(\alpha\) is a contrast parameter; in our experiments, we set \(\alpha=1\), making it the direct average of matched similarities. Further details on generation and scoring appear in App.~\ref{app:persona-generation-details} and App.~\ref{app:persona-score-details}, respectively.

\subsubsection{Structural Features and Operation Tags}

To characterize task structure, we parse the gold SQL with SQLGlot
\citep{sqlglot2026} into an eight-dimensional feature vector comprising counts of distinct columns, tables, \texttt{WHERE} predicates, joins, aggregates, scalar functions, common table expressions, and nested subqueries.
These features follow the query-structure profiling of \citet{maEvaluatingLlmsTexttosql2024}.
We include the full vector with the released dataset and report its equally-weighted sum as the \gls{scs}.

\subsubsection{Schema Generation}

We evaluate three schema representations per database to compare raw DDL, data-profiled schema context, and LLM-enriched schema context:
\begin{itemize}[leftmargin=*,topsep=0pt,itemsep=1pt]
    \item \textit{DDL}: raw SQL \texttt{CREATE TABLE} statements from \base{}, exposing table and column names, declared types, and constraints.
    \item \textit{Schema}: a cognitive-catalog representation with row and column counts, column types, inferred roles, null and uniqueness rates, summary statistics, bounded example values, and available BIRD column descriptions, but no LLM-generated descriptions.
    \item \textit{SchemaLLM}: the same catalog structure, supplemented with GPT-5.2-generated column descriptions and table summaries where original dictionaries are incomplete.
\end{itemize}
Additional details are in App.~\ref{app:schema-variant-details}.

\vspace{\SubsectionSpacing}
\subsection{Mixed-Effects Modeling}
\vspace{\SubsectionSpacing}
\label{subsec:evaluation-protocol}

Each evaluation row is one generated answer for a task, \gls{llm}, target paradigm, persona condition, and schema representation. Because tasks repeat across prompt and schema conditions, and tasks from the same database share schema and domain context, we model execution accuracy with mixed-effects logistic regressions rather than treating rows as independent trials. Models are fit separately for each LLM so the estimated SQL-pandas gap and moderators are model-specific.

The outcome is binary semantic correctness. We execute generated SQL against the SQLite database and generated pandas against the corresponding CSV tables, then compare each output to the gold output using the \base{} semantic-equivalence protocol \citep{huBenchmarkingTexttoPythonTexttoSQL2026}. Every row also records the metadata needed to audit and stratify the results, including the generated code, execution output, judge output, errors, and timing.

The fixed effects encode the experimental factors. With SQL and Original DDL as reference levels for the categorical predictors, the Python coefficient estimates the pandas-SQL difference under the reference schema at mean-centered covariates. PS and SCS are standardized continuous covariates, making their effects comparable on the log-odds scale.

We fit two models that differ only in how they represent task structure. Both
share the same outcome, random-effects structure, and a common set of fixed
effects: paradigm, standardized PS, schema representation, and the Python
interaction with each. The primary model adds the aggregate SCS and its Python
interaction, testing whether structurally demanding tasks are harder overall and
whether that burden is paradigm-specific.
The focused model instead replaces SCS with four operation-level predictors chosen to probe structural patterns that plausibly differ between declarative SQL and procedural pandas: subquery presence, selected-column count, scalar-function presence, and join count, each with a Python interaction.
We log-one-plus transform and standardize the selected-column and join counts to reduce skew from high-count queries, and treat subquery and scalar-function use as binary indicators, since their nonzero counts are sparse.

The random effects absorb the dependence induced by the matched design. A
question-level intercept lets some tasks be intrinsically easier or harder
across all conditions, a question-level paradigm slope lets the pandas penalty
vary by task, and a database-level intercept captures shared schema and domain
effects among questions from the same database. Full details are in
App.~\ref{app:mixed-effects-specifications}.
\vspace{\SectionSpacing}
\section{Experiments and Results}
\vspace{\SectionSpacing}
\label{sec:artifact-analyses}

We first report artifact-level diagnostics for \name{}, then turn to model-based SQL-pandas analysis. The diagnostics establish persona separation and structural coverage; the mixed-effects analyses use these annotations to analyze execution accuracy across target paradigms.

\vspace{\SubsectionSpacing}
\subsection{Experiment Setup}
\vspace{\SubsectionSpacing}

We evaluate nine LLMs: Gemini-2.5-Flash and Gemini-2.5-Pro \citep{googleGeminiModels2026}, Gemma-4-26B \citep{googleGemma4HF2026}, GPT-5-Mini and GPT-5.1 \citep{openaiModels2026}, Qwen3-14B and Qwen3-8B \citep{qwen3technicalreport}, Qwen3-Coder-30B \citep{qwen3coder30bHF2025}, and GLM-4-32B \citep{zaiGLM432BHF2025}. For each model, we run the full cross-product of two target paradigms (SQL and pandas), three persona conditions (original, novice, and expert), and three schema representations (Original DDL, Schema, and SchemaLLM).
Seventeen samples from the base \base{} dataset were missing gold SQL labels, and one sample was missing the query; we excluded these samples from the analysis.
The resulting prediction-evaluation matrix contains 245,592 rows, or 27,288 evaluated predictions per model or 13,644 per model-paradigm pair. Generation is run with deterministic decoding where supported. We evaluate generations with the \base{} semantic-equivalence protocol, yielding the binary \texttt{is\_correct} outcome analyzed below.

\vspace{\SubsectionSpacing}
\subsection{Persona Quality Audit}
\vspace{\SubsectionSpacing}

The PS confirms that the expert rewrites are substantially more
schema-explicit than the novice rewrites. Across all tasks, the mean
schema-specificity score is 0.567 for novice questions and 0.960 for expert
questions, with a mean expert-minus-novice gap of 0.393.
The original \base{} questions score
0.635 on average-closer to the novice end in 89.0\% of tasks.
We provide examples of persona-stratified queries in App. \ref{app:persona_examples}.

As an audit of PS, we measured pairwise overlap among the original, novice, and expert questions (Table~\ref{tab:lexical-diversity-summary}) to verify that the persona conditions were not minor rewrites. We reported symmetric sentence-level BLEU-4 and BLEU-2, plus lowercased whitespace-token Jaccard overlap to capture order-invariant vocabulary sharing. As expected, original-novice pairs overlapped most, reflecting everyday phrasings of the same request; pairs involving the expert rewrite overlapped less because they added schema names, joins, filters, and output constraints. This pattern confirmed the intended separation between business-style and schema-explicit phrasings. Additional word-count analyses are provided in App.~\ref{app:person_word_count}.

\begin{table}[t]
\centering
\scriptsize
\setlength{\tabcolsep}{4pt}
\resizebox{\columnwidth}{!}{%
\begin{tabular}{lrrr}
\toprule
\textbf{Condition pair} & \textbf{BLEU-4 $\uparrow$} & \textbf{BLEU-2 $\uparrow$} & \textbf{Jaccard $\uparrow$} \\
\midrule
Original-novice & 0.108 & 0.254 & 0.307 \\
Original-expert & 0.008 & 0.025 & 0.078 \\
Novice-expert & 0.005 & 0.018 & 0.066 \\
\bottomrule
\end{tabular}%
}
    \vspace{-2mm}
    \caption{Mean pairwise overlap metrics. BLEU scores are symmetric sentence-BLEU averages. Lower values indicate greater lexical divergence.}
    \vspace{-2mm}
\label{tab:lexical-diversity-summary}
\end{table}

The audit identified 39 tasks with an expert-minus-novice gap below 0.2 (23 below 0.1); these are not necessarily invalid since short SQL with few extractable tokens yields uninformative scores, and all such rows are retained with their observed scores for explicit filtering.

A complementary blind human audit of 250 tasks found that all 150
representative-random pairs retained the reference task and that all 150
schema-explicit variants in this stratum contained only gold-supported
constraints. All 40 schema-explicit variants in the targeted output-contract
stratum were also gold-supported (App.~\ref{app:human-audit-protocol}).

\vspace{\SubsectionSpacing}
\subsection{Structural and Operation Coverage}
\vspace{\SubsectionSpacing}

Table~\ref{tab:structural-feature-summary} shows the per-component and composite
score distributions. The composite structural score spans a wide range (2-50,
$\sigma{=}4.8$) with a median of 11 and an interquartile range of 8-13,
indicating sufficient variance for quartile-based stratification despite the
BIRD-derived complexity floor. Columns and scalar functions exhibit the highest
per-component variance ($\sigma{=}2.08$ and $1.76$), making them the primary
drivers of task-level score differences.

\begin{table}[t]
\centering
\small
\setlength{\tabcolsep}{4pt}
\begin{tabular}{lrrrr}
\toprule
\textbf{Feature} & \textbf{Mean} & \textbf{Median} & \textbf{Max} & \textbf{Std} \\
\midrule
Columns & 4.71 & 5 & 30 & 2.08 \\
Tables & 1.96 & 2 & 5 & 0.72 \\
\texttt{WHERE} predicates & 1.54 & 1 & 8 & 0.95 \\
Join pairs & 0.96 & 1 & 6 & 0.79 \\
Aggregations & 0.61 & 0 & 5 & 0.78 \\
Scalar functions & 1.06 & 1 & 23 & 1.76 \\
CTEs & 0.02 & 0 & 3 & 0.18 \\
Subqueries & 0.13 & 0 & 4 & 0.42 \\
\midrule
\textit{SCS} & \textit{10.99} & \textit{11} & \textit{50} & \textit{4.80} \\
\bottomrule
\end{tabular}
    \vspace{-2mm}
    \caption{A summary of SQL-derived structural features. The SCS is the equal-weighted sum of all features.}
    \vspace{-2mm}
\label{tab:structural-feature-summary}
\end{table}

Table~\ref{tab:dataset-difficulty-summary} profiles the benchmark by database
and \base{} difficulty level. The average structural score increases from
simple to moderate to challenging tasks in every database, which supports using
the continuous SQL-derived score as a finer-grained covariate while retaining
\base{}'s coarse difficulty labels for descriptive stratification.

\begin{table}[t]
\centering
\small
\setlength{\tabcolsep}{4pt}
\resizebox{\columnwidth}{!}{%
\begin{tabular}{lcccc}
\toprule
\textbf{Database} & \textbf{Simple} & \textbf{Moderate} & \textbf{Challenging} & \textbf{Overall} \\
\midrule
california\_schools & 10 & 12 & 16 & 11 \\
card\_games & 8 & 12 & 13 & 9 \\
codebase\_community & 8 & 11 & 14 & 9 \\
debit\_card\_specializing & 9 & 15 & 20 & 11 \\
european\_football\_2 & 9 & 12 & 16 & 11 \\
financial & 13 & 17 & 22 & 15 \\
formula\_1 & 10 & 15 & 24 & 13 \\
student\_club & 9 & 13 & 15 & 10 \\
superhero & 9 & 14 & 15 & 11 \\
thrombosis\_prediction & 10 & 12 & 16 & 12 \\
toxicology & 9 & 12 & 11 & 10 \\
\bottomrule
\end{tabular}%
}
    \vspace{-2mm}
    \caption{Average structural score for each database grouped by BIRD-Python difficulty level.}
    \vspace{-2mm}
\label{tab:dataset-difficulty-summary}
\end{table}

\vspace{\SubsectionSpacing}
\subsection{Dataset Release}
\vspace{\SubsectionSpacing}

The release retains all 1,534 source rows while marking analysis eligibility.
The 18 samples missing gold SQL labels or queries are included in the dataset for visibility but excluded from all analyses.
The release is available at \url{https://huggingface.co/datasets/optum-cognition-lab/BIRD-Persona}.

\subsection{Model-Based SQL-pandas Analysis}

We next compare SQL and pandas execution accuracy using descriptive aggregates and two mixed-effects models.

Table~\ref{tab:execution-accuracy-main} reports raw execution accuracy by model
and output paradigm. SQL exceeds pandas for 8 of the 9 models, with gaps
from 0.034 (GLM-4-32B) to 0.210 (Qwen3-Coder-30B). Gemma-4-26B is the only model
with higher pandas than SQL accuracy.

\begin{table}[t]
\centering
\small
\setlength{\tabcolsep}{4pt}
\begin{tabular}{lcc}
\toprule
\textbf{Model} & \textbf{pandas} & \textbf{SQL} \\
\midrule
Gemini-2.5-Flash & 0.674 {\scriptsize (0.004)} & 0.741 {\scriptsize (0.004)} \\
Gemini-2.5-Pro & 0.714 {\scriptsize (0.004)} & 0.765 {\scriptsize (0.004)} \\
Gemma-4-26B & 0.666 {\scriptsize (0.004)} & 0.623 {\scriptsize (0.004)} \\
GPT-5-Mini & 0.612 {\scriptsize (0.004)} & 0.714 {\scriptsize (0.004)} \\
GPT-5.1 & 0.634 {\scriptsize (0.004)} & 0.721 {\scriptsize (0.004)} \\
Qwen3-14B & 0.500 {\scriptsize (0.004)} & 0.690 {\scriptsize (0.004)} \\
Qwen3-8B & 0.574 {\scriptsize (0.004)} & 0.690 {\scriptsize (0.004)} \\
Qwen3-Coder-30B & 0.477 {\scriptsize (0.004)} & 0.688 {\scriptsize (0.004)} \\
GLM-4-32B & 0.587 {\scriptsize (0.004)} & 0.621 {\scriptsize (0.004)} \\
\bottomrule
\end{tabular}%
    \vspace{-2mm}
    \caption{Execution accuracy by model and output paradigm, aggregated over all
persona conditions and schema variants. Parentheses report binomial standard
errors for the binary correctness indicator.}
\vspace{-2mm}
\label{tab:execution-accuracy-main}
\end{table}

Table~\ref{tab:evaluation-condition-summary} summarizes the same evaluation
matrix by persona condition, schema representation, and binned SCS. Persona
produces the largest descriptive shift: expert prompts reach 0.889 SQL and 0.765
pandas accuracy, while novice prompts fall to 0.561 and 0.494. Schema
representation has a much smaller effect, consistent with the model-dependent
schema results discussed below. Accuracy declines monotonically with SCS in both
paradigms. Full model-by-persona and model-by-schema tables appear in
App.~\ref{app:prediction-evaluation-tables}.

\begin{table}[t]
\centering
\small
\setlength{\tabcolsep}{4pt}
\resizebox{\columnwidth}{!}{%
\begin{tabular}{llcc}
\toprule
\textbf{Factor} & \textbf{Condition} & \textbf{pandas} & \textbf{SQL} \\
\midrule
\multirow{3}{*}{Persona}
  & Expert & 0.766 {\scriptsize (0.002)} & 0.889 {\scriptsize (0.002)} \\
  & Novice & 0.494 {\scriptsize (0.002)} & 0.561 {\scriptsize (0.002)} \\
  & Original & 0.553 {\scriptsize (0.002)} & 0.634 {\scriptsize (0.002)} \\
\midrule
\multirow{3}{*}{Schema}
  & Original DDL & 0.608 {\scriptsize (0.002)} & 0.704 {\scriptsize (0.002)} \\
  & Schema & 0.602 {\scriptsize (0.002)} & 0.691 {\scriptsize (0.002)} \\
  & SchemaLLM & 0.604 {\scriptsize (0.002)} & 0.689 {\scriptsize (0.002)} \\
\midrule
\multirow{3}{*}{\shortstack[l]{SCS (Binned)}}
  & Low (0-10) & 0.678 {\scriptsize (0.002)} & 0.754 {\scriptsize (0.002)} \\
  & Medium (11-20) & 0.563 {\scriptsize (0.002)} & 0.666 {\scriptsize (0.002)} \\
  & High (21+) & 0.338 {\scriptsize (0.007)} & 0.468 {\scriptsize (0.007)} \\
\bottomrule
\end{tabular}%
}
    \vspace{-2mm}
    \caption{Execution accuracy by persona, schema representation,
    and binned SCS. Each row averages over models and all experimental factors
    other than the condition shown, separately by paradigm. Parentheses report
    binomial standard errors for the binary correctness indicator.}
    \vspace{-2mm}
\label{tab:evaluation-condition-summary}
\end{table}

We next fit the mixed-effects models summarized in
Section~\ref{subsec:evaluation-protocol}, with primary-model fixed effects in
Table~\ref{tab:model1-results-main}. After adjustment, the SQL advantage
persists: the pandas coefficient is negative for 8 of 9 systems, from
$-0.65$ (GLM-4-32B) to $-2.25$ (Qwen3-Coder-30B). Gemma-4-26B is again the
exception, with an indistinguishable coefficient ($+0.01$, $p=0.87$). The aggregate gap therefore persists after adjustment for PS, structure, schema representation, and question/database heterogeneity.

\begin{table*}[t]
\centering
\small
\setlength{\tabcolsep}{4pt}
\resizebox{\textwidth}{!}{%
\begin{tabular}{lrrrrrrrrr}
\toprule
\multicolumn{1}{c}{} & \multicolumn{5}{c}{\emph{Main Effects}} & \multicolumn{4}{c}{\emph{Interaction Effects}} \\
\cmidrule(lr){2-6}\cmidrule(lr){7-10}
\textbf{Model} & \shortstack{\textbf{pandas}} & \shortstack{\textbf{PS}} & \shortstack{\textbf{SCS}} & \shortstack{\textbf{Schema}} & \shortstack{\textbf{SchemaLLM}} & \shortstack{\textbf{pandas $\times$}\\\textbf{PS}} & \shortstack{\textbf{pandas $\times$}\\\textbf{SCS}} & \shortstack{\textbf{pandas $\times$}\\\textbf{Schema}} & \shortstack{\textbf{pandas $\times$}\\\textbf{SchemaLLM}} \\
\midrule
Gemini-2.5-Flash & \underline{\textbf{-1.10}} & \underline{\textbf{1.84}} & \underline{\textbf{-0.54}} & \underline{\textbf{0.32}} & \underline{\textbf{0.24}} & \underline{\textbf{-0.82}} & -0.07 & -0.16 & -0.13 \\
Gemini-2.5-Pro & \underline{\textbf{-1.39}} & \underline{\textbf{2.02}} & \underline{\textbf{-0.53}} & 0.05 & -0.02 & \underline{\textbf{-0.63}} & 0.03 & \underline{\textbf{0.30}} & \underline{\textbf{0.37}} \\
Gemma-4-26B & 0.02 & \underline{\textbf{0.88}} & \underline{\textbf{-0.50}} & \underline{\textbf{-0.52}} & \underline{\textbf{-0.47}} & \underline{\textbf{0.11}} & -0.06 & \underline{\textbf{0.72}} & \underline{\textbf{0.71}} \\
GPT-5-Mini & \underline{\textbf{-1.34}} & \underline{\textbf{2.22}} & -0.13 & 0.07 & -0.06 & \underline{\textbf{-0.52}} & -0.10 & \textbf{-0.22} & -0.08 \\
GPT-5.1 & \underline{\textbf{-1.51}} & \underline{\textbf{2.20}} & \underline{\textbf{-0.32}} & 0.06 & 0.01 & \underline{\textbf{-0.67}} & 0.12 & 0.08 & 0.11 \\
Qwen3-14B & \underline{\textbf{-1.82}} & \underline{\textbf{1.73}} & \underline{\textbf{-0.40}} & -0.08 & -0.03 & \underline{\textbf{-0.89}} & \underline{\textbf{-0.38}} & -0.08 & -0.11 \\
Qwen3-8B & \underline{\textbf{-1.23}} & \underline{\textbf{1.62}} & \underline{\textbf{-0.47}} & -0.11 & -0.12 & \underline{\textbf{-0.90}} & \underline{\textbf{-0.24}} & -0.17 & -0.17 \\
Qwen3-Coder-30B & \underline{\textbf{-2.25}} & \underline{\textbf{1.78}} & \underline{\textbf{-0.48}} & \underline{\textbf{-0.23}} & \textbf{-0.17} & \underline{\textbf{-0.96}} & \underline{\textbf{-0.32}} & 0.00 & 0.03 \\
GLM-4-32B & \underline{\textbf{-0.65}} & \underline{\textbf{1.68}} & \underline{\textbf{-0.50}} & \underline{\textbf{-0.45}} & \underline{\textbf{-0.53}} & \underline{\textbf{-0.59}} & -0.06 & \textbf{0.19} & \underline{\textbf{0.35}} \\
\bottomrule
\end{tabular}%
}
\vspace{-2mm}
\caption{Fixed-effect coefficients for the primary mixed-effects logistic regression model. Statistically significant results are \textbf{Bolded} ($\alpha = 0.05$) or \underline{\textbf{Underlined}} ($\alpha=0.01$) after FDR correction. $p$-values are reported in App. \ref{app:main_model_results}.}
\label{tab:model1-results-main}
\vspace{-2mm}
\end{table*}

The primary model also reports the covariate effects. PS is positive for every
model, and the pandas $\times$ PS interaction is negative for eight models
(Gemma-4-26B is the only positive interaction). Aggregate SCS is negative for
all models and significant for most, while the pandas $\times$ SCS interaction
is significant only for the three Qwen systems. Schema effects are less stable:
both the main effects and pandas $\times$ schema interactions vary in sign and
significance across models, matching the weak schema shifts in
Table~\ref{tab:evaluation-condition-summary}. Full coefficients, standard
errors, and raw $p$-values are in App.~\ref{app:main_model_results}.

In an additional control for the raw number of gold table, column, and literal
tokens named in the question, PS remained positive and FDR-significant in all
nine models. Likelihood-ratio tests likewise showed that the PS terms improved
fit beyond the raw token count in all nine models
(App.~\ref{app:schema-token-control}).

The focused decomposition model splits aggregate complexity into
operation-specific predictors (full table in
App.~\ref{app:complexity_model_results}). The pandas $\times$ log column-count
coefficient is negative for 8 of 9 systems, the exception being GPT-5.1.
The subquery main effect is negative for all models, while the pandas $\times$
subquery interaction is positive for all models. Join count and
scalar-function presence show no comparable modality-specific pattern.
Table~\ref{tab:model2-results-main} reports the corresponding structural
interaction coefficients.

\begin{table}[t]
\centering
\scriptsize
\setlength{\tabcolsep}{2pt}
\resizebox{\columnwidth}{!}{%
\begin{tabular}{lrrrr}
\toprule
\textbf{Model} & \shortstack{\textbf{pandas $\times$}\\\textbf{\texttt{SUBQUERY}}} & \shortstack{\textbf{pandas $\times$}\\\textbf{\texttt{COLUMN}}} & \shortstack{\textbf{pandas $\times$}\\\textbf{\texttt{SCALAR}}} & \shortstack{\textbf{pandas $\times$}\\\textbf{\texttt{JOIN}}} \\
\midrule
Gemini-2.5-Flash & \underline{\textbf{1.37}} & -0.18 & 0.10 & -0.01 \\
Gemini-2.5-Pro & \underline{\textbf{0.71}} & -0.11 & 0.29 & -0.07 \\
Gemma-4-26B & \underline{\textbf{2.39}} & \underline{\textbf{-0.52}} & -0.23 & \underline{\textbf{0.39}} \\
GPT-5-Mini & 0.18 & \textbf{-0.23} & 0.22 & -0.02 \\
GPT-5.1 & \underline{\textbf{0.85}} & 0.08 & 0.26 & \textbf{-0.29} \\
Qwen3-14B & \underline{\textbf{1.05}} & \underline{\textbf{-0.53}} & -0.13 & -0.06 \\
Qwen3-8B & \underline{\textbf{0.82}} & \underline{\textbf{-0.31}} & 0.00 & -0.08 \\
Qwen3-Coder-30B & \underline{\textbf{0.83}} & \underline{\textbf{-0.52}} & \textbf{0.35} & -0.09 \\
GLM-4-32B & \underline{\textbf{1.15}} & \underline{\textbf{-0.49}} & \underline{\textbf{0.54}} & 0.12 \\
\bottomrule
\end{tabular}%
}
\vspace{-2mm}
\caption{Selected interaction-effect coefficients from the structural decomposition model (full results in App.~\ref{app:complexity_model_results}). Subquery and scalar predictors are binary presence indicators; column and join predictors are standardized log counts. FDR-significant results are \textbf{bold} ($\alpha{=}0.05$) or \underline{\textbf{underlined}} ($\alpha{=}0.01$).}
\label{tab:model2-results-main}
\vspace{-2mm}
\end{table}

\vspace{\SectionSpacing}
\section{Discussion}
\vspace{\SectionSpacing}
\label{sec:discussion}

In this paper, we introduced \name{} as a benchmark for evaluating SQL and pandas generation under variation in user phrasing, schema representation, and task structure.

\ParagraphHeaderSpacing
\textbf{The pandas target remains less reliable than SQL for most systems.}
Across models, \name{} reproduces the SQL-over-Python pattern reported in prior cross-paradigm work: 8 out of 9 systems show lower pandas accuracy than SQL after accounting for phrasing specificity, schema representation, task structure, and repeated task/database effects. This suggests that target paradigm is not just a formatting choice. Generating pandas code requires models to sequence dataframe operations, manage intermediate objects, and preserve output shape, making it a deployment-relevant evaluation dimension. Gemma-4-26B is the sole exception, with higher pandas than SQL accuracy. A deeper analysis suggests this may be driven by a comparatively high SQL execution-error rate (16.4\% vs.\ 9.8\% for pandas; Table~\ref{tab:execution-error-rate}), where the most common failures are syntax/parse errors and references to nonexistent columns. This suggests that Gemma's lower SQL accuracy partly reflects difficulty producing executable SQL rather than an inability to solve the underlying task, though we avoid a stronger causal claim absent a dedicated ablation.

\ParagraphHeaderSpacing
\textbf{User phrasing is the central benchmark axis.}
Phrasing specificity is the most consistent predictor of correctness. Schema-explicit prompts improve both SQL and pandas, but the gain is larger for SQL, where named tables, columns, predicates, and aggregation map directly onto target clauses. For user-facing analytics systems, business-style prompts are therefore not merely shorter technical prompts; they create a different reliability profile, especially when the target is pandas. This association persists even after controlling for the raw number of gold identifiers named in the prompt: PS remains a positive, significant predictor in all nine models (App.~\ref{app:schema-token-control}), indicating that the effect is not driven solely by schema-token overlap.

\ParagraphHeaderSpacing
\textbf{Schema enrichment is model-dependent.}
Schema variants produce mixed, model-dependent effects rather than a consistent improvement. One possible explanation is that many \base{} schemas have interpretable table and column names, so additional descriptions add little marginal information or introduce prompt noise; private or enterprise schemas with opaque abbreviations may behave differently. 

\ParagraphHeaderSpacing
\textbf{Task structure explains where the gap appears.}
Aggregate structural complexity summarizes overall difficulty, but operation-level annotations better localize target-specific failures. Selected-column burden widens the pandas gap, plausibly because pandas code must explicitly select, name, and preserve output columns across intermediate steps. Subqueries, by contrast, are difficult overall but not uniquely difficult for pandas, since nested declarative structure can sometimes be decomposed into sequential dataframe transformations. This shows why aggregate accuracy alone is insufficient for diagnosing paradigm reliability.

\ParagraphHeaderSpacing
\textbf{Implications for user-facing evaluation.}
\name{} suggests evaluating SQL and pandas systems under real-world conditions: varying phrasing styles, available schema context at inference, and expected task structures. The benchmark can help to guide prompt and context design by identifying where clarification, schema grounding, or output-shape constraints are most needed.
\vspace{\SectionSpacing}
\section{Conclusion}
\vspace{\SectionSpacing}
\label{sec:conclusion}

We introduced \name{}, a persona-stratified extension of \base{} for matched SQL and pandas analytics code generation. Across nine LLMs, the benchmark reproduces the previously reported SQL-over-Python gap while showing that the gap varies with user phrasing, schema representation, and task structure. These results argue for evaluation protocols that go beyond aggregate accuracy and test user-facing analytics systems under the user, system, and task conditions in which each target paradigm is expected to operate.
\section*{Limitations}
This study has limitations at both the artifact and analysis levels. First, the persona paraphrase pipeline generates and filters rewrites automatically. The PS is useful for constructing schema-explicitness contrasts, but it measures schema grounding rather than full semantic fidelity to the gold task. We supplement these automated checks with a blind human audit of 250 paper eligible task triples-a representative random sample of 150 and 100 targeted cases using final adjudicated labels (App.~\ref{app:human-audit-protocol}). Although the audit provides evidence of task preservation and gold-supported explicit constraints, it does not constitute exhaustive validation of all 1,534 tasks. The expert and novice conditions also differ not only in linguistic register but also in length and information content: expert rewrites average roughly 58 words versus 14 for novice and explicitly name schema identifiers, filters, and output constraints. Our expert-versus-novice contrast is therefore defined by schema explicitness, and the observed persona effects should not be attributed to register alone; isolating a pure register effect would require a schema-fixed, length-matched ablation that varies only phrasing while holding information content constant, which we leave to future work.

Second, all generated persona variants come from a single generator, which may imprint model-specific lexical and stylistic regularities on the novice and expert conditions. The generator is excluded from the evaluated systems, but we cannot fully rule out stylistic overlap with the downstream model families. Relatedly, the SchemaLLM descriptions are generated by GPT-5.2, which shares a model family with two evaluated systems (GPT-5.1 and GPT-5-Mini), raising the possibility of a within-family advantage. Empirically, we see no such pattern: SchemaLLM is not statistically significant for the within-family models. However, our design does not formally test this effect.

Third, the base tasks come from the publicly released BIRD development set, so possible exposure to the underlying schemas, SQL, or original questions cannot be ruled out for some evaluated models. This caveat applies mainly to the inherited base corpus rather than the newly generated persona rewrites and schema variants; it may also help explain why added schema context yields only limited gains.

Fourth, the structural representation is deliberately simple and therefore partial. It captures major aspects of relational query structure, including joins, aggregations, selected columns, scalar functions, and subqueries, but does not characterize every source of task difficulty, and the operation tags are rule-based AST flags that do not yet cover all semantic operations.

In addition, we did not conduct a controlled scaling study and therefore cannot draw conclusions about the relationship between model size and the SQL-pandas gap. Several evaluated models are closed-weight with undisclosed parameter counts; even among open-weight models, size is confounded with architecture, training data, instruction tuning, and inference design. We leave a controlled model-scale analysis for future work.

Finally, our explanatory analyses are observational. The interaction patterns across user phrasing, paradigm, and task structure are therefore best interpreted as associations indicating when the SQL-pandas gap widens or narrows, not as proof of a single causal mechanism.

\bibliography{dh_references,other_sources}

\clearpage
\appendix

\section{Background}

\subsection{Related Work} \label{app:background:related_works}
Below we review related work on \tts{} evaluation, \ttp{} and cross-paradigm
code generation, persona variability, schema representation, and structural
complexity.

\ParagraphHeaderSpacing
\textbf{Realistic \tts{} evaluation.}
\tts{} benchmarks have moved from controlled cross-domain generalization toward
larger, more realistic database settings. Spider established the standard
multi-database benchmark for schema generalization \citep{yu2018spider}, and
BIRD added realistic database-grounded questions and execution
\citep{li2023bird}. Spider 2.0 shifts further toward enterprise analyst
workflows with harder schemas and more realistic tasks \citep{lei2025spider2},
and UNITE offers a broader unified suite for \tts{} evaluation
\citep{lan2023unite}. We build on this line but focus on matched SQL-pandas
comparison and persona-stratified variation rather than expanding \tts{}
coverage alone.

\ParagraphHeaderSpacing
\textbf{\ttp{} and cross-paradigm evaluation.}
Recent work has begun to evaluate data-analysis code generation beyond SQL.
DSCodeBench studies realistic data-science problems in Python with executable
tests across common libraries \citep{ouyang2026dscodebench}. Closest to our
setting is \base{}, which aligns each question with both SQL and pandas
solutions, allowing \tts{} and \ttp{} to be compared on the same analytics task
\citep{huBenchmarkingTexttoPythonTexttoSQL2026}. We extend this setting with
controlled persona rewrites, schema variants, and task-level covariates for
analyzing when the paradigms diverge.

\ParagraphHeaderSpacing
\textbf{User variability, ambiguity, and robustness.}
Several lines of work show that user phrasing, expertise, and
underspecification can substantially affect evaluation. \citet{palta-etal-2025-speaking}
study five levels of user expertise in general LLM interactions, but do not
evaluate code generation tasks or matched novice and expert versions of the same
task. In \tts{}, AmbiQT and AmbiSQL make ambiguity explicit and show how
multiple plausible interpretations complicate single-gold evaluation
\citep{bhaskar2023ambiqt,ding2025ambisql}. SEDE captures naturally
underspecified questions from Stack Exchange \citep{hazoom2021sede}, while work
on healthcare \tts{} highlights the effect of domain terminology,
abbreviations, and typos \citep{wang2020emr}. Spider-Syn studies robustness to
meaning-preserving synonym substitutions \citep{gan2021spidersyn}. These works
motivate our persona axis: novice and expert phrasings preserve the target
answer while changing how much schema and operational detail is explicit.

\ParagraphHeaderSpacing
\textbf{Schema linking and schema representation.}
Schema information remains a central design variable in \tts{} systems, even as
stronger LLMs complicate earlier assumptions about explicit schema linking.
Recent work studies whether schema linking remains necessary
\citep{maamari2024death}, improves linking through bidirectional or contextual
retrieval \citep{cao2024rslsql,nahid2026rethinking}, and scales linking to large
multi-database settings \citep{wang2025linkalign}. CHESS provides a
retrieval-heavy pipeline that uses contextual filtering for SQL synthesis
\citep{talaei2024chess}, while enterprise-oriented analysis argues that NL2SQL
remains difficult under realistic schema complexity and ambiguity
\citep{floratou2024nl2sql}. Our benchmark treats schema representation as an
experimental condition and tests whether schema information changes SQL and
pandas accuracy differently.

\ParagraphHeaderSpacing
\textbf{Complexity-aware \tts{} analysis.}
Prior work also shows that aggregate benchmark scores can hide important
variation across query structure. Ma, Pu, and Zhu compare TPC-DS, BIRD, and
Spider through structural complexity features that capture columns, tables,
predicates, joins, aggregates, scalar functions, common table expressions, and
subqueries \citep{ma2025complexqueries}. Related work on SQL-complexity
measurement for LLM evaluation motivates treating query structure as a
continuous covariate rather than only a coarse difficulty label
\citep{maEvaluatingLlmsTexttosql2024}. Beyond SELECT further argues that
realistic \tts{} evaluation should cover a wider range of SQL operations than
SELECT-heavy benchmarks \citep{wang2025beyondselect}. We adopt this
complexity-aware view to analyze whether structural features affect \tts{} and
\ttp{} differently.

\section{Method \& Experiment Details}

\subsection{Artifact Generation}

\subsubsection{Schema Variant Construction}
\label{app:schema-variant-details}

Schema variants are generated once per database and reused across all questions
from that database. The construction uses the BIRD-Python SQLite databases,
flat CSV exports, and database-description files. The current schema-variant
manifest covers 11 databases, 75 tables, and 798 columns. Of the 798 original
dictionary rows, 652 contain non-empty source column descriptions and 146 are
missing source descriptions.

The \texttt{original\_ddl} variant is written directly from SQLite metadata by
extracting the \texttt{CREATE TABLE} statements for each non-system table. The
two catalog variants are built by running the cognitive catalog pipeline over
the flat CSV tables and then rendering a text catalog for prompting. The
\texttt{cognitive\_with\_descriptions} variant keeps only the source-provided
BIRD column descriptions. The
\texttt{cognitive\_with\_llm\_descriptions} variant keeps the same catalog
shape but uses the cognitive engine's native LLM-facing catalog text, which
includes LLM-filled missing column descriptions and LLM-generated table
descriptions.

\subsubsection{Structural Feature Definition Details}

Our structural-complexity representation is an eight-dimensional feature vector
computed from the gold SQL AST. The dimensions are: number of distinct column
references, number of physical tables, number of atomic \texttt{WHERE}
predicates, number of join pairs, number of aggregate expressions, number of
non-aggregate scalar function calls, number of common table expressions, and
number of nested subqueries. The scalar \texttt{task\_structural\_score} used in the main paper is the equal-weight sum of these counts.

\subsubsection{Persona Rewrite Generation}
\label{app:persona-generation-details}

For each \base{} task, GPT-5.2 received the original question and gold SQL. The
implementation supports optional schema context, but it was not supplied in the
full-dataset generation run. The database identifier is used only
for logging. It returns a business-style/underspecified rewrite and a
schema-explicit rewrite as a JSON object. The prompt controls four dimensions:
schema and scope, values and formats, logic and business rules, and output
formatting and constraints. It instructs the generator to keep both rewrites
anchored to the source task (entity, measure, population, and goal), to make
expert-only explicit details gold-supported, and to let the business-style
rewrite omit detail without introducing a contradictory constraint. Thus, these
are generation constraints rather than a claim that every rewrite alone uniquely
determines the gold executable query.

After each parseable response, we compute the expert-minus-novice PS gap. We
accept a pair when the gap is at least 0.2; otherwise, we append feedback that
lists up to 15 extracted SQL tokens, asks the schema-explicit rewrite to state
them and its supported values and output constraints, and asks the business-style
rewrite to avoid schema-specific terms. We make up to three scored attempts. If
no SQL tokens can be extracted, we accept the first parseable response because
the gap cannot be evaluated. After exhausted attempts, the pipeline makes one
final best-effort call with the latest feedback; if it cannot produce parseable
JSON, the task is recorded as a generation failure. The released artifact retains
the final questions, scores, and gap, including pairs below the threshold. Of
the 1,534 released pairs, 39 remain below the 0.2 threshold after generation;
they are retained and identifiable from \texttt{gap} $< 0.2$.
\paragraph{Prompt specification.}
The GPT-5.2 system prompt requested two complete, grammatical JSON fields:
\texttt{expert} and \texttt{novice}. It specified the four controls above.
For the schema-explicit field, it required exact table and column names, gold
literal values, and all gold-supported aggregation, tie-breaking, null-handling,
ordering, limit, and output-column constraints; joins and computations had to be
described in natural language rather than transcribed as SQL. For the
business-style field, it prohibited verbatim schema identifiers and literal SQL
values and requested business vocabulary with implicit logic and presentation
details. Both fields were instructed to retain the source task's entity, measure,
population, and goal; the business-style field could omit detail but could not
introduce a new filter, condition, or constraint.

The exact versioned system prompt and retry-feedback template will accompany
the released source artifact.

\subsubsection{Persona-Score Details}
\label{app:persona-score-details}

The Persona Score (PS) measures this phrasing specificity, the explicitness of
schema, value, and operator mentions, rather than semantic equivalence. Given a gold SQL query and a question \(q\), we extract schema and
value tokens \(T\) from the SQL, including table names, column names, string
literals, date strings, and numeric constants. SQL parsing is performed with
SQLGlot, using the SQLite dialect first and a dialect-unspecified parse as a
fallback; the scorer walks the SQL abstract syntax tree and records table nodes,
non-\texttt{*} column nodes, and literal nodes. If parsing fails, a regex
fallback extracts quoted string literals, backtick identifiers, numeric
literals, and non-keyword word tokens. Extracted tokens are normalized by
lowercasing, stripping quotes, and replacing identifier separators such as
underscores with spaces, then deduplicated after filtering SQL keywords and
tokens shorter than two non-space characters.

For each question, we construct candidate spans \(C(q)\) from unigrams, bigrams,
and trigrams over lowercased regex word tokens. Punctuation separates words,
while underscores are retained by the tokenizer; therefore a query mention such
as \texttt{customer\_id} is a high character-level match to the normalized SQL
token \textit{customer id}, but is not forced to be an exact match. Similarity
between a SQL token \(t \in T\) and candidate span \(c \in C(q)\) is the
character-level Ratcliff-Obershelp ratio,
\[
  s(t,c)=\frac{2m(t,c)}{|t|+|c|},
\]
where \(m(t,c)\) is the total length of non-overlapping matching character
blocks. This lexical score intentionally rewards exact or near-exact schema
mentions rather than semantic synonyms; for example, a question that says
\textit{client} should not receive the same schema-specificity credit as one
that names \texttt{customer\_id}.

To avoid double-counting, we greedily sort all token-candidate pairs by
similarity and accept pairs in descending order only when neither the SQL token
nor any word in the candidate span has already been used. This one-to-one
constraint
prevents a single vague word such as \textit{customer} from
satisfying several schema tokens such as \texttt{customer\_id},
\texttt{customer\_name}, and \texttt{customer\_type}. The accepted ratio for
SQL token \(t\) is denoted \(r_t \in [0,1]\), with \(r_t=1\) for an exact
lexical match and \(r_t=0\) for unmatched tokens. When \(T\) is nonempty, the
final score is the mean \(\alpha\)-transformed matched similarity over extracted
SQL tokens,
\[
  \mathrm{PS}(q,\mathrm{SQL})
  = \frac{1}{|T|}\sum_{t\in T} r_t^{\alpha}.
\]
The contrast parameter \(\alpha\) controls how strongly partial lexical matches
are penalized. The experiments use \(\alpha=1\), which leaves the raw
Ratcliff-Obershelp ratios unchanged; larger values would increasingly penalize
partial matches while preserving exact matches.

For a paraphrase pair, we compute
\(\mathrm{gap}=\mathrm{PS}(q_{\mathrm{expert}},\mathrm{SQL})-
\mathrm{PS}(q_{\mathrm{novice}},\mathrm{SQL})\). The generation loop compares
this gap with a configurable target. If the gap is smaller, feedback is appended
to the prompt asking the expert rewrite to include concrete schema identifiers
and the novice rewrite to avoid them; regeneration is attempted up to three
times. The final dataset preserves the original, novice, and expert scores and
the expert-minus-novice gap for downstream auditing, including rows whose final
gap remains small after retry.

\subsubsection{Human Validation and Semantic Preservation}
\label{app:human-audit-protocol}

We conducted a single-blind audit of 250 paper-eligible task triples. The sample
contains a representative random stratum of 150 tasks and 100 targeted tasks:
40 with output contracts (ordering, limits, or output shape), 25 with filters
or thresholds, 20 with aggregation or grouping, and 15 with joins or subqueries.
The targeted strata are reported separately rather than treated as representative
of the corpus.

Each task was assessed with the variants presented as A/B in randomized order.
The blinded review included the original request, both variants, the full gold
SQL, and a deterministic SQL-constraint summary, but not the persona labels,
model predictions, or evaluation outcomes.

For each variant, the audit checked whether every explicit detail was supported
by the gold SQL, whether the request contradicted the gold target, and whether
the request and schema uniquely determined the gold output. It separately
checked whether the pair retained the reference task's entity, measure,
population, and goal. Omitting a gold detail in a business-style rewrite was not
itself treated as a contradiction.

All 250 audited pairs retained the reference task, including all 150 pairs in
the representative random stratum. Overall, 246 of 250 expert schema-explicit variants
had all explicit constraints supported by the gold SQL; all 40 expert schema-explicit
variants in the targeted output-contract stratum were gold-supported.
These results assess task anchoring and support of stated details rather than
assuming that every novice business-style request alone uniquely determines the gold
executable query. In eight novice business-style variants from the random stratum,
a gold-output constraint could not be inferred from the prompt alone.
These cases were adjudicated as underdetermined rather than contradictory: the
prompt did not state a constraint that conflicted with the gold target,
and the shared reference task was preserved. Thus, missing an inferable
output constraint was treated separately from changing the task or introducing
an unsupported requirement.

\subsection{Release Eligibility}
\label{subsec:eligibility}

The source artifact has 1,534 unique question identifiers. Seventeen rows have
empty gold SQL in the local BIRD-Python file and are quarantined from
SQL-derived structural analyses. One row has an empty original question and is
quarantined from original-question evaluation until repaired. Structural
analyses therefore use the 1,517 rows with non-empty SQL and successful SQL
parses. The dataset release preserves the quarantined rows for
traceability but exposes \texttt{analysis\_eligible} and
\texttt{quarantine\_reason} fields so that benchmark consumers can reproduce the
main denominators.

\subsection{Dataset Fields}

The released table exposes the following fields for every source task:
\texttt{question\_id}, \texttt{db\_id}, \texttt{difficulty},
\texttt{question}, \texttt{novice\_question}, \texttt{expert\_question},
\texttt{sql}, \texttt{code}, \texttt{evidence},
\texttt{original\_similarity\_score},
\texttt{novice\_similarity\_score},
\texttt{expert\_similarity\_score}, \texttt{gap},
\texttt{structural\_features}, \texttt{extended\_constructs},
\texttt{task\_structural\_score}, \texttt{operation\_tags},
\texttt{analysis\_eligible}, and \texttt{quarantine\_reason}. The eligibility
fields are required because the source artifact intentionally preserves rows
that are excluded from one or more analyses.

\subsubsection{Evaluation Row Schema}

Each prediction row includes \texttt{question\_id}, \texttt{db\_id},
\texttt{difficulty}, \texttt{persona\_condition}, \texttt{question\_field},
\texttt{paradigm}, \texttt{model\_id}, \texttt{code},
\texttt{input\_tokens}, \texttt{output\_tokens}, and \texttt{elapsed\_sec}.
Each evaluation row should additionally include \texttt{is\_correct},
\texttt{actual\_output}, \texttt{expected\_output},
\texttt{judge\_actual\_output}, \texttt{judge\_expected\_output},
\texttt{llm\_response}, and \texttt{error}. This schema is intentionally
redundant so that aggregate tables can be reproduced without reconstructing
metadata from separate files.

\subsection{Mixed-Effects Model Specifications}
\label{app:mixed-effects-specifications}

This appendix gives the full statistical specification used for the reported
explanatory analyses. All reported regressions are fit separately for each
evaluated LLM. The unit of observation is one model output for a fixed task,
target paradigm, persona condition, and schema condition. The response variable
is binary semantic correctness after execution-based evaluation:
\[
  y_i =
  \begin{cases}
    1, & \text{if output } i \text{ is semantically correct},\\
    0, & \text{otherwise.}
  \end{cases}
\]
We use a binomial generalized linear mixed model with a logit link. For each
observation $i$, let $q(i)$ denote its question identifier and $d(i)$ denote its
database identifier. SQL is the reference paradigm, and original DDL is the
reference schema representation. The pandas indicator is denoted $P_i$, where
$P_i=1$ for pandas and $P_i=0$ for SQL.

\subsubsection{Shared Preprocessing}

We first restrict the modeling data to complete cases for the response,
question identifier, database identifier, persona score, and all covariates
included in the corresponding model. Before fitting these models, we also apply
the analysis exclusion filter from the notebook, removing 17 question identifiers
with missing or unverifiable ground-truth labels. This leaves 1,517 source tasks
and 27,306 observations in each LLM-specific model fit. The paradigm factor is
encoded with SQL as the reference level:
\[
  \texttt{paradigm} \in \{\texttt{sql}, \texttt{pandas}\}.
\]
Schema representation is encoded as a three-level factor with levels DDL,
Schema, and SchemaLLM. DDL is the reference level. The two non-reference schema
indicators are denoted $S^{\mathrm{Schema}}_i$ and
$S^{\mathrm{SchemaLLM}}_i$.

Persona specificity is computed from the overlap between the question text and
SQL-grounded schema and literal tokens, as described in
Section~\ref{sec:methods}. For each observation, we use the score associated
with that observation's persona condition. The score is standardized within the
analysis data:
\[
  Z^{\mathrm{persona}}_i =
  \frac{\mathrm{persona\_score}_i -
  \overline{\mathrm{persona\_score}}}
  {\mathrm{sd}(\mathrm{persona\_score})}.
\]
All continuous structural predictors are standardized after any transformation
specified below. Thus, continuous coefficients correspond to a one-standard-
deviation increase in the transformed predictor.

\subsubsection{Primary Aggregate-Complexity Model}
\label{app:primary-model-spec}

The primary model estimates the overall SQL-pandas gap and how that gap changes
with persona specificity, aggregate structural complexity, and schema
representation. The aggregate structural score is the equal-weight sum of the
eight SQL-derived structural feature counts:
\[
  \begin{aligned}
  C_i =&\ c^{\mathrm{col}}_i + c^{\mathrm{tab}}_i
  + c^{\mathrm{where}}_i + c^{\mathrm{join}}_i \\
  &+ c^{\mathrm{agg}}_i + c^{\mathrm{scalar}}_i
  + c^{\mathrm{cte}}_i + c^{\mathrm{subq}}_i .
  \end{aligned}
\]
We standardize this score before modeling:
\[
  Z^{\mathrm{complexity}}_i =
  \frac{C_i-\overline{C}}{\mathrm{sd}(C)}.
\]
For compactness, let $Z^A_i$ denote standardized persona specificity and let
$Z^C_i$ denote standardized aggregate complexity. The primary fixed-effect
linear predictor is

\begingroup
\small
\[
\begin{aligned}
  \eta_i
  =&\ \beta_0
  + \beta_P P_i
  + \beta_A Z^A_i
  + \beta_C Z^C_i \\
  &+ \beta_D S^{\mathrm{Schema}}_i
  + \beta_L S^{\mathrm{SchemaLLM}}_i \\
  &+ \beta_{PA} P_i Z^A_i
  + \beta_{PC} P_i Z^C_i \\
  &+ \beta_{PD} P_i S^{\mathrm{Schema}}_i
  + \beta_{PL} P_i S^{\mathrm{SchemaLLM}}_i
  \\
  &+ b_{0,q(i)}
  + b_{P,q(i)} P_i
  + u_{0,d(i)} .
\end{aligned}
\]
\endgroup

The response model is
\[
  y_i \sim \mathrm{Bernoulli}(\pi_i),
  \qquad
  \mathrm{logit}(\pi_i)=\eta_i .
\]
The question-level random effects allow each question to have its own baseline
difficulty and its own pandas-vs-SQL shift:
\[
  \begin{pmatrix}
    b_{0,q} \\
    b_{P,q}
  \end{pmatrix}
  \sim
  \mathcal{N}
  \left(
    \begin{pmatrix}0\\0\end{pmatrix},
    \Sigma_q
  \right).
\]
The database-level random intercept is
\[
  u_{0,d} \sim \mathcal{N}(0,\sigma^2_d).
\]
In \texttt{lme4} notation, the model is:
\begin{quote}
\small
\begin{verbatim}
correct_semantic ~
  paradigm * z_persona_score +
  paradigm * z_complexity +
  paradigm * schema_variant +
  (1 + paradigm | question_id) +
  (1 | db_id)
\end{verbatim}
\end{quote}
The coefficient $\beta_P$ estimates the pandas-vs-SQL difference at mean
persona specificity, mean aggregate complexity, and original DDL schema.
Interaction coefficients such as $\beta_{PA}$ and $\beta_{PC}$ estimate how the
pandas-vs-SQL gap changes as persona specificity or aggregate complexity
increases.

\subsubsection{Focused Structural Decomposition Model}
\label{app:structural-decomposition-model}

The focused structural decomposition model replaces the aggregate complexity
score with a small set of operation-specific predictors chosen to test
modality-specific procedural burden. The included structural dimensions are
subquery presence, selected-column count, scalar-function presence, and join
count. These dimensions capture nested decomposition, output projection burden,
local transformation burden, and relational composition, respectively. Common
table expressions are excluded because they occur in only 1.1\% of questions,
making model-based estimates unstable and likely to be driven by a small number
of items. \texttt{WHERE} predicate count, aggregate count, and physical table
count are omitted from the focused model to keep the confirmatory structural
interaction family small and theoretically targeted; the aggregate-complexity
model already captures overall task difficulty.

The analysis notebook also fits a broader seven-feature structural
decomposition as an exploratory intermediate model. The reported structural
tables use the focused specification below.

For selected columns and join pairs, we apply a log-one-plus transformation and
then standardize:
\[
  Z_{ik}^{\log} =
  \frac{\log(1+c_{ik}) -
  \overline{\log(1+c_k)}}
  {\mathrm{sd}(\log(1+c_k))}.
\]
The \(\log(1+x)\) transform preserves the ordering and zero values of the
original count variables while reducing leverage from rare high-count examples.
After standardization, the column-count and join-count coefficients are
interpreted as effects of a one-standard-deviation increase in the
log-transformed count. Subqueries and scalar functions are encoded as binary
presence indicators:
\begin{align*}
  H^{\mathrm{subq}}_i
  &= \mathbbm{1}\{c^{\mathrm{subq}}_i > 0\},\\
  H^{\mathrm{scalar}}_i
  &= \mathbbm{1}\{c^{\mathrm{scalar}}_i > 0\}.
\end{align*}
The binary encoding is used for these dimensions because the scientific
question is whether the operation type is present, and because the nonzero
count distributions are sparse relative to selected columns and joins.

Let $Z^{\log,\mathrm{col}}_i$ denote standardized log selected-column count and
let $Z^{\log,\mathrm{join}}_i$ denote standardized log join-pair count. The
focused structural decomposition linear predictor is
\begingroup
\small
\[
\begin{aligned}
  \eta_i
  =&\ \beta_0
  + \beta_P P_i
  + \beta_A Z^A_i \\
  &+ \beta_Q H^{\mathrm{subq}}_i
  + \beta_M Z^{\log,\mathrm{col}}_i
  \\
  &+ \beta_F H^{\mathrm{scalar}}_i
  + \beta_J Z^{\log,\mathrm{join}}_i \\
  &+ \beta_D S^{\mathrm{Schema}}_i
  + \beta_L S^{\mathrm{SchemaLLM}}_i \\
  &+ \beta_{PA} P_i Z^A_i \\
  &+ \beta_{PQ} P_i H^{\mathrm{subq}}_i \\
  &+ \beta_{PM} P_i Z^{\log,\mathrm{col}}_i \\
  &+ \beta_{PF} P_i H^{\mathrm{scalar}}_i \\
  &+ \beta_{PJ} P_i Z^{\log,\mathrm{join}}_i \\
  &+ \beta_{PD} P_i S^{\mathrm{Schema}}_i
  + \beta_{PL} P_i S^{\mathrm{SchemaLLM}}_i \\
  &+ b_{0,q(i)}
  + b_{P,q(i)} P_i
  + u_{0,d(i)} .
\end{aligned}
\]
\endgroup
The response distribution and random-effect distributions are the same as in
Appendix~\ref{app:primary-model-spec}. In \texttt{lme4} notation, the model is:
\begin{quote}
\small
\begin{verbatim}
correct_semantic ~
  paradigm * z_persona_score +
  paradigm * has_subquery +
  paradigm * z_log_column_count +
  paradigm * has_scalar_function +
  paradigm * z_log_join_count +
  paradigm * schema_variant +
  (1 + paradigm | question_id) +
  (1 | db_id)
\end{verbatim}
\end{quote}
For a structural feature $k$, the corresponding main-effect coefficient is the
effect of that feature for SQL, the reference paradigm. The interaction with
$P_i$ is the additional effect for pandas relative to SQL. Therefore, the total
pandas-side effect of that feature is the sum of the main effect and the
pandas interaction. This decomposition is the basis for statements about which
structural operations widen or narrow the SQL-pandas gap.

\subsubsection{Estimation and Reporting}

The two reported model families are estimated with maximum likelihood using
the Laplace approximation as implemented in \texttt{lme4::glmer}. We use a
binomial family with a logit link and the \texttt{bobyqa} optimizer with
\texttt{maxfun = 2e5}. Each model is fit independently for each LLM, so fixed
effects are interpreted within model rather than as pooled cross-model
contrasts. We report fixed-effect estimates on the log-odds scale, with
standard errors, Wald $z$ statistics, and Wald $p$-values. We also report model
fit summaries, including AIC, BIC, log likelihood, AUROC, and the number of
observations used in each fit.

For multiplicity-aware interpretation, the fixed-effect output files also
include Benjamini-Hochberg false-discovery-rate thresholds and significance
indicators. The correction family is defined conservatively within each model
table as all non-intercept fixed-effect tests across the nine LLM-specific
fits. Intercepts are excluded because the null hypothesis that the baseline
log-odds equals zero is not a scientific hypothesis of interest. Full result
tables report raw Wald $p$-values, while the summary tables use the corrected
indicators for significance formatting after accounting for multiple
comparisons.

The random-effect term
\((1 + \texttt{paradigm} \mid \texttt{question\_id})\) is included because the
SQL-pandas gap can vary by item. The random intercept captures question-level
baseline difficulty, while the random pandas slope captures question-specific
deviation from the average pandas-vs-SQL effect. The term
\((1 \mid \texttt{db\_id})\) captures residual database-level heterogeneity.
These random effects are not nuisance terms only: their variance components are
used to assess whether the modality gap is a global offset or varies
substantially across questions.

\section{Additional Results}

\subsection{Persona Word-Count Audit} \label{app:person_word_count}

A word-count analysis (Table \ref{tab:persona-quality-summary}) reveals that expert rewrites are nearly four
times longer than original questions (mean 57.6 vs.\ 14.5 words), reflecting
the additional schema references, filter specifications, and output constraints
they encode. Novice questions average 14.3 words, only a 0.2 word difference from the original query, suggesting that the persona manipulation changes specificity rather
than verbosity at the business-level register.

\begin{table}[t]
\centering
\small
\setlength{\tabcolsep}{4pt}
\resizebox{\columnwidth}{!}{%
\begin{tabular}{lrrrr}
\toprule
\textbf{Condition} & \textbf{Mean score} & \textbf{Median score} & \textbf{Mean words} & \textbf{Median words} \\
\midrule
Original & 0.635 & 0.630 & 14.5 & 14.0 \\
Novice & 0.567 & 0.570 & 14.3 & 14.0 \\
Expert & 0.960 & 0.971 & 57.6 & 56.0 \\
\bottomrule
\end{tabular}%
}
    \vspace{-2mm}
    \caption{Persona quality audit.
    }
\label{tab:persona-quality-summary}
\end{table}

\subsection{Persona-Score Example}
\label{app:persona-score-example}

Consider a gold query whose extracted tokens include
\texttt{customer\_id}, \texttt{order\_id}, \texttt{total\_amount},
\texttt{order\_date}, \texttt{orders}, \texttt{customer\_name}, and date
literals. After SQL-token normalization, these become spans such as
\textit{customer id}, \textit{order id}, and \textit{total amount}. A novice
question such as \textit{Show me each customer's biggest recent order} contains
useful business words, but it does not name most schema identifiers. Without
one-to-one matching, the word \textit{customer} could partially satisfy both
\textit{customer id} and \textit{customer name}, and the word \textit{order}
could partially satisfy \textit{order id}, \textit{order date}, and
\textit{orders}. The greedy matching step prevents this inflation by allowing
each question word to contribute to at most one SQL token.

Under the greedy assignment, the novice phrasing receives high credit for only a
small number of root-word overlaps, while missing dates and specific columns
contribute zero or near-zero scores. An expert phrasing such as
\textit{For each customer\_id, return the order\_id with max total\_amount where
order\_date is in 2024, sorted by customer\_name} directly names the schema
fields and therefore receives high per-token ratios for most identifiers. In
the worked example from the pipeline documentation, the novice matched ratios
average to roughly 0.33, while the expert ratios average to roughly 0.69,
yielding an expert-minus-novice gap of about 0.36. If a generated pair falls
below the configured gap target, the pipeline asks the generator to make the
expert rewrite more schema-specific and the novice rewrite less
schema-specific; after retries, the final artifact retains the observed scores
so low-gap rows can be audited rather than silently discarded.

\subsection{Execution-Error Analysis}
\label{app:execution-error-analysis}

Table~\ref{tab:execution-error-rate} reports execution-error rates by model and output paradigm. Gemma-4-26B is the only model whose SQL error rate (16.4\%) exceeds its pandas error rate (9.8\%); this 6.6-point difference is larger than its 4.3-point pandas accuracy advantage.
Inspecting Gemma's recorded SQL errors shows the most common categories are syntax/parse errors and references to nonexistent columns, suggesting the lower SQL accuracy is driven in part by difficulty producing executable SQL rather than necessarily reflecting an inability to solve the underlying data-analysis task.

\begin{table}[t]
\centering
\small
\setlength{\tabcolsep}{4pt}
\begin{tabular}{lcc}
\toprule
\textbf{Model} & \textbf{SQL error rate} & \textbf{pandas error rate} \\
\midrule
Gemini-2.5-Pro   & 0.3\%  & 3.6\%  \\
Gemini-2.5-Flash & 1.7\%  & 7.0\%  \\
GPT-5.1          & 1.9\%  & 4.0\%  \\
GPT-5-Mini       & 1.9\%  & 2.2\%  \\
Qwen3-14B        & 4.2\%  & 19.7\% \\
Qwen3-8B         & 5.0\%  & 19.1\% \\
Qwen3-Coder-30B  & 5.4\%  & 20.5\% \\
GLM-4-32B        & 10.5\% & 12.9\% \\
Gemma-4-26B      & 16.4\% & 9.8\%  \\
\bottomrule
\end{tabular}%
\vspace{-2mm}
\caption{Execution-error rates by model and output paradigm, aggregated over all
persona conditions and schema variants. A generation counts as an execution
error when it raises a runtime error (an SQLite error for SQL, or a Python
exception for pandas). Gemma-4-26B is the only model whose SQL error rate
exceeds its pandas error rate, consistent with its outlier pandas-over-SQL
accuracy in Table~\ref{tab:execution-accuracy-main}.}
\vspace{-2mm}
\label{tab:execution-error-rate}
\end{table}

\subsection{Prediction Evaluation Tables}
\label{app:prediction-evaluation-tables}

Tables~\ref{tab:evaluation-by-persona-model} and
\ref{tab:evaluation-by-schema-model} expand the main raw evaluation summary.
These tables report descriptive execution accuracy and are not adjusted for
question- or database-level dependence. They are included to make the evaluation
matrix auditable before the mixed-effects analyses.

\begin{table*}[t]
\centering
\small
\setlength{\tabcolsep}{4pt}
\begin{tabular}{lcccccc}
\toprule
& \multicolumn{2}{c}{\textbf{Expert}} & \multicolumn{2}{c}{\textbf{Novice}} & \multicolumn{2}{c}{\textbf{Original}} \\
\cmidrule(lr){2-3}\cmidrule(lr){4-5}\cmidrule(lr){6-7}
\textbf{Model} & \textbf{pandas} & \textbf{SQL} & \textbf{pandas} & \textbf{SQL} & \textbf{pandas} & \textbf{SQL} \\
\midrule
Gemini-2.5-Flash & 0.803 {\scriptsize (0.006)} & 0.915 {\scriptsize (0.004)} & 0.576 {\scriptsize (0.007)} & 0.616 {\scriptsize (0.007)} & 0.641 {\scriptsize (0.007)} & 0.692 {\scriptsize (0.007)} \\
Gemini-2.5-Pro & 0.856 {\scriptsize (0.005)} & 0.924 {\scriptsize (0.004)} & 0.608 {\scriptsize (0.007)} & 0.647 {\scriptsize (0.007)} & 0.679 {\scriptsize (0.007)} & 0.722 {\scriptsize (0.007)} \\
Gemma-4-26B & 0.809 {\scriptsize (0.006)} & 0.766 {\scriptsize (0.006)} & 0.565 {\scriptsize (0.007)} & 0.520 {\scriptsize (0.007)} & 0.623 {\scriptsize (0.007)} & 0.581 {\scriptsize (0.007)} \\
GPT-5-Mini & 0.843 {\scriptsize (0.005)} & 0.928 {\scriptsize (0.004)} & 0.454 {\scriptsize (0.007)} & 0.568 {\scriptsize (0.007)} & 0.541 {\scriptsize (0.007)} & 0.645 {\scriptsize (0.007)} \\
GPT-5.1 & 0.831 {\scriptsize (0.006)} & 0.927 {\scriptsize (0.004)} & 0.503 {\scriptsize (0.007)} & 0.575 {\scriptsize (0.007)} & 0.569 {\scriptsize (0.007)} & 0.661 {\scriptsize (0.007)} \\
Qwen3-14B & 0.658 {\scriptsize (0.007)} & 0.900 {\scriptsize (0.004)} & 0.395 {\scriptsize (0.007)} & 0.544 {\scriptsize (0.007)} & 0.449 {\scriptsize (0.007)} & 0.626 {\scriptsize (0.007)} \\
Qwen3-8B & 0.694 {\scriptsize (0.007)} & 0.887 {\scriptsize (0.005)} & 0.494 {\scriptsize (0.007)} & 0.556 {\scriptsize (0.007)} & 0.534 {\scriptsize (0.007)} & 0.626 {\scriptsize (0.007)} \\
Qwen3-Coder-30B & 0.640 {\scriptsize (0.007)} & 0.897 {\scriptsize (0.004)} & 0.376 {\scriptsize (0.007)} & 0.546 {\scriptsize (0.007)} & 0.416 {\scriptsize (0.007)} & 0.620 {\scriptsize (0.007)} \\
GLM-4-32B & 0.757 {\scriptsize (0.006)} & 0.855 {\scriptsize (0.005)} & 0.476 {\scriptsize (0.007)} & 0.476 {\scriptsize (0.007)} & 0.528 {\scriptsize (0.007)} & 0.533 {\scriptsize (0.007)} \\
\bottomrule
\end{tabular}%
    \vspace{-2mm}
    \caption{Execution accuracy by model, persona condition, and output paradigm,
aggregated over schema variants, with standard errors in parentheses.}
\label{tab:evaluation-by-persona-model}
\end{table*}

\begin{table*}[t]
\centering
\small
\setlength{\tabcolsep}{4pt}
\begin{tabular}{lcccccc}
\toprule
& \multicolumn{2}{c}{\textbf{DDL (Baseline)}} & \multicolumn{2}{c}{\textbf{Schema}} & \multicolumn{2}{c}{\textbf{SchemaLLM}} \\
\cmidrule(lr){2-3}\cmidrule(lr){4-5}\cmidrule(lr){6-7}
\textbf{Model} & \textbf{pandas} & \textbf{SQL} & \textbf{pandas} & \textbf{SQL} & \textbf{pandas} & \textbf{SQL} \\
\midrule
Gemini-2.5-Flash & 0.664 {\scriptsize (0.007)} & 0.726 {\scriptsize (0.007)} & 0.681 {\scriptsize (0.007)} & 0.752 {\scriptsize (0.006)} & 0.675 {\scriptsize (0.007)} & 0.745 {\scriptsize (0.006)} \\
Gemini-2.5-Pro & 0.694 {\scriptsize (0.007)} & 0.764 {\scriptsize (0.006)} & 0.725 {\scriptsize (0.007)} & 0.767 {\scriptsize (0.006)} & 0.725 {\scriptsize (0.007)} & 0.763 {\scriptsize (0.006)} \\
Gemma-4-26B & 0.649 {\scriptsize (0.007)} & 0.666 {\scriptsize (0.007)} & 0.672 {\scriptsize (0.007)} & 0.598 {\scriptsize (0.007)} & 0.677 {\scriptsize (0.007)} & 0.604 {\scriptsize (0.007)} \\
GPT-5-Mini & 0.622 {\scriptsize (0.007)} & 0.714 {\scriptsize (0.007)} & 0.607 {\scriptsize (0.007)} & 0.719 {\scriptsize (0.007)} & 0.607 {\scriptsize (0.007)} & 0.708 {\scriptsize (0.007)} \\
GPT-5.1 & 0.625 {\scriptsize (0.007)} & 0.720 {\scriptsize (0.007)} & 0.640 {\scriptsize (0.007)} & 0.724 {\scriptsize (0.007)} & 0.637 {\scriptsize (0.007)} & 0.720 {\scriptsize (0.007)} \\
Qwen3-14B & 0.515 {\scriptsize (0.007)} & 0.694 {\scriptsize (0.007)} & 0.492 {\scriptsize (0.007)} & 0.685 {\scriptsize (0.007)} & 0.494 {\scriptsize (0.007)} & 0.690 {\scriptsize (0.007)} \\
Qwen3-8B & 0.600 {\scriptsize (0.007)} & 0.698 {\scriptsize (0.007)} & 0.562 {\scriptsize (0.007)} & 0.686 {\scriptsize (0.007)} & 0.560 {\scriptsize (0.007)} & 0.685 {\scriptsize (0.007)} \\
Qwen3-Coder-30B & 0.496 {\scriptsize (0.007)} & 0.701 {\scriptsize (0.007)} & 0.462 {\scriptsize (0.007)} & 0.679 {\scriptsize (0.007)} & 0.474 {\scriptsize (0.007)} & 0.684 {\scriptsize (0.007)} \\
GLM-4-32B & 0.605 {\scriptsize (0.007)} & 0.657 {\scriptsize (0.007)} & 0.572 {\scriptsize (0.007)} & 0.608 {\scriptsize (0.007)} & 0.583 {\scriptsize (0.007)} & 0.599 {\scriptsize (0.007)} \\
\bottomrule
\end{tabular}%
    \vspace{-2mm}
    \caption{Execution accuracy by model, schema representation, and output
paradigm, aggregated over persona conditions, with standard errors in parentheses.}
\label{tab:evaluation-by-schema-model}
\end{table*}

\subsection{Detailed Logistic Regression Model Results: Main Model} \label{app:main_model_results}

Table~\ref{tab:model1-results-full} reports the full fixed-effect
coefficient table for the primary mixed-effects logistic regression, estimated
separately for each of the nine LLMs. Predictors are on the logit scale.

\begin{table*}[!ht]
    \centering
    \small
    \resizebox{\textwidth}{!}{%
    \begin{tabular}{llrrrrrrrrrr}
    \toprule
        \textbf{Values} & \textbf{model\_id} & \textbf{(Intercept)} & \textbf{pandas} & \shortstack{\textbf{pandas $\times$}\\\textbf{Schema}} & \shortstack{\textbf{pandas $\times$}\\\textbf{SchemaLLM}} & \shortstack{\textbf{pandas $\times$}\\\textbf{SCS}} & \shortstack{\textbf{pandas $\times$}\\\textbf{PS}} & \textbf{Schema} & \textbf{SchemaLLM} & \textbf{SCS} & \textbf{PS} \\
    \midrule
        \multirow{9}{*}{Coefficient} & Gemini-2.5-Flash & 2.68 & -1.1 & -0.158 & -0.13 & -0.0727 & -0.825 & 0.321 & 0.239 & -0.54 & 1.84 \\
         & Gemini-2.5-Pro & 3.61 & -1.39 & 0.298 & 0.366 & 0.0252 & -0.635 & 0.0499 & -0.0155 & -0.525 & 2.02 \\
         & Gemma-4-26B & 1.32 & 0.0168 & 0.721 & 0.71 & -0.0571 & 0.111 & -0.52 & -0.466 & -0.503 & 0.877 \\
         & GPT-5-Mini & 2.6 & -1.34 & -0.219 & -0.0787 & -0.0958 & -0.524 & 0.0747 & -0.0635 & -0.133 & 2.22 \\
         & GPT-5.1 & 2.73 & -1.51 & 0.0827 & 0.106 & 0.118 & -0.673 & 0.0604 & 0.00821 & -0.315 & 2.2 \\
         & Qwen3-14B & 1.94 & -1.82 & -0.0789 & -0.109 & -0.377 & -0.887 & -0.0802 & -0.0348 & -0.401 & 1.73 \\
         & Qwen3-8B & 2.01 & -1.23 & -0.17 & -0.17 & -0.238 & -0.9 & -0.112 & -0.123 & -0.471 & 1.62 \\
         & Qwen3-Coder-30B & 2.25 & -2.25 & -0.000456 & 0.0254 & -0.319 & -0.965 & -0.228 & -0.173 & -0.476 & 1.78 \\
         & GLM-4-32B & 1.57 & -0.65 & 0.192 & 0.354 & -0.0632 & -0.593 & -0.455 & -0.529 & -0.503 & 1.68 \\
    \midrule
        \multirow{9}{*}{Standard Error} & Gemini-2.5-Flash & 0.352 & 0.0885 & 0.0949 & 0.0946 & 0.0668 & 0.0529 & 0.072 & 0.0717 & 0.094 & 0.0456 \\
         & Gemini-2.5-Pro & 0.371 & 0.0959 & 0.104 & 0.104 & 0.0697 & 0.0595 & 0.0781 & 0.0779 & 0.109 & 0.0512 \\
         & Gemma-4-26B & 0.256 & 0.0776 & 0.082 & 0.0821 & 0.0609 & 0.0396 & 0.0561 & 0.0561 & 0.06 & 0.0271 \\
         & GPT-5-Mini & 0.312 & 0.0799 & 0.0955 & 0.0953 & 0.0552 & 0.0532 & 0.0722 & 0.0719 & 0.098 & 0.0481 \\
         & GPT-5.1 & 0.281 & 0.0869 & 0.096 & 0.0959 & 0.063 & 0.0555 & 0.0734 & 0.0732 & 0.0999 & 0.0491 \\
         & Qwen3-14B & 0.228 & 0.0835 & 0.0838 & 0.0839 & 0.0695 & 0.046 & 0.0645 & 0.0646 & 0.0752 & 0.0397 \\
         & Qwen3-8B & 0.241 & 0.0781 & 0.085 & 0.085 & 0.0605 & 0.0443 & 0.065 & 0.065 & 0.0768 & 0.0379 \\
         & Qwen3-Coder-30B & 0.207 & 0.0904 & 0.0854 & 0.0854 & 0.0763 & 0.0471 & 0.0673 & 0.0674 & 0.0846 & 0.0413 \\
         & GLM-4-32B & 0.327 & 0.0792 & 0.0852 & 0.0852 & 0.0619 & 0.0442 & 0.063 & 0.063 & 0.0744 & 0.0357 \\
    \midrule
        \multirow{9}{*}{$p$-value} & Gemini-2.5-Flash & 2.48e-14 & 1.16e-35 & 0.0951 & 0.168 & 0.277 & 8.1e-55 & 8.38e-6 & 0.000855 & 9.14e-9 & 0 \\
         & Gemini-2.5-Pro & 2.06e-22 & 9.11e-48 & 0.00407 & 0.000413 & 0.718 & 1.45e-26 & 0.523 & 0.842 & 1.33e-6 & 0 \\
         & Gemma-4-26B & 2.39e-7 & 0.829 & 1.52e-18 & 5.63e-18 & 0.349 & 0.00528 & 1.82e-20 & 9.56e-17 & 4.91e-17 & 2.24e-230 \\
         & GPT-5-Mini & 7.16e-17 & 5.59e-63 & 0.0219 & 0.409 & 0.0829 & 6.81e-23 & 0.301 & 0.377 & 0.176 & 0 \\
         & GPT-5.1 & 2.48e-22 & 1.36e-67 & 0.389 & 0.269 & 0.0619 & 7.17e-34 & 0.411 & 0.911 & 0.00159 & 0 \\
         & Qwen3-14B & 1.64e-17 & 8.05e-105 & 0.346 & 0.194 & 5.75e-8 & 8.02e-83 & 0.214 & 0.59 & 9.38e-8 & 0 \\
         & Qwen3-8B & 6.11e-17 & 3.18e-56 & 0.0455 & 0.0452 & 8.13e-5 & 6.07e-92 & 0.0847 & 0.0584 & 8.54e-10 & 0 \\
         & Qwen3-Coder-30B & 1.32e-27 & 3.25e-136 & 0.996 & 0.766 & 2.93e-5 & 3.3e-93 & 0.000718 & 0.0105 & 1.79e-8 & 0 \\
         & GLM-4-32B & 1.6e-6 & 2.26e-16 & 0.0244 & 3.32e-5 & 0.308 & 4.49e-41 & 5.15e-13 & 4.14e-17 & 1.37e-11 & 0 \\
    \bottomrule
    \end{tabular}%
    }
    \vspace{-2mm}
    \caption{Full fixed-effect results for the primary mixed-effects logistic regression, estimated separately for each LLM. Logit-scale coefficients, standard errors, and $p$-values are reported in separate row groups.}
    \label{tab:model1-results-full}
\end{table*}

Table~\ref{tab:model1-fit} reports the model fit summaries for the primary
mixed-effects logistic regression, including AIC, BIC, log likelihood, and
AUROC for each LLM-specific fit.

\begin{table*}[t]
\centering
\scriptsize
\setlength{\tabcolsep}{4pt}
\begin{tabular}{lrrrr}
\toprule
\textbf{Model} & \textbf{AIC} & \textbf{BIC} & \textbf{Log Likelihood} & \textbf{AUROC} \\
\midrule
Gemini-2.5-Flash & 20988.64 & 21103.64 & -10480.32 & 0.947 \\
Gemini-2.5-Pro & 18366.91 & 18481.91 & -9169.45 & 0.960 \\
Gemma-4-26B & 25770.22 & 25885.22 & -12871.11 & 0.918 \\
GPT-5-Mini & 20927.93 & 21042.93 & -10449.97 & 0.951 \\
GPT-5.1 & 21018.84 & 21133.84 & -10495.42 & 0.951 \\
Qwen3-14B & 25589.43 & 25704.43 & -12780.71 & 0.927 \\
Qwen3-8B & 24813.13 & 24928.13 & -12392.56 & 0.926 \\
Qwen3-Coder-30B & 25469.39 & 25584.39 & -12720.70 & 0.928 \\
GLM-4-32B & 24796.81 & 24911.81 & -12384.40 & 0.933 \\
\bottomrule
\end{tabular}%
    \vspace{-2mm}
    \caption{Model fit summaries for the primary mixed-effects logistic regression, estimated separately for each LLM.}
\label{tab:model1-fit}
\end{table*}

Table~\ref{tab:model1-random-effects} reports the random-effect summaries for
the primary mixed-effects logistic regression, including query-level and
database-level variance components for each LLM-specific fit.

\begin{table*}[t]
\centering
\small
\setlength{\tabcolsep}{4pt}
\begin{tabular}{lllccc}
\toprule
\textbf{Model} & \textbf{Group} & \textbf{Component} & \textbf{Variance} & \textbf{SD} & \textbf{Correlation} \\
\midrule
\multirow{4}{*}{Gemini-2.5-Flash} & \multirow{3}{*}{Query} & Intercept & 10.26 & 3.20 & - \\
 &  & pandas & 3.67 & 1.91 & - \\
 &  & Intercept-pandas Covariance & - & - & -0.597 \\
 & Database & Intercept & 1.24 & 1.12 & - \\
\midrule
\multirow{4}{*}{Gemini-2.5-Pro} & \multirow{3}{*}{Query} & Intercept & 13.58 & 3.68 & - \\
 &  & pandas & 3.59 & 1.90 & - \\
 &  & Intercept-pandas Covariance & - & - & -0.541 \\
 & Database & Intercept & 1.35 & 1.16 & - \\
\midrule
\multirow{4}{*}{Gemma-4-26B} & \multirow{3}{*}{Query} & Intercept & 3.90 & 1.98 & - \\
 &  & pandas & 3.38 & 1.84 & - \\
 &  & Intercept-pandas Covariance & - & - & -0.257 \\
 & Database & Intercept & 0.671 & 0.819 & - \\
\midrule
\multirow{4}{*}{GPT-5-Mini} & \multirow{3}{*}{Query} & Intercept & 11.54 & 3.40 & - \\
 &  & pandas & 1.80 & 1.34 & - \\
 &  & Intercept-pandas Covariance & - & - & -0.639 \\
 & Database & Intercept & 0.941 & 0.970 & - \\
\midrule
\multirow{4}{*}{GPT-5.1} & \multirow{3}{*}{Query} & Intercept & 12.00 & 3.46 & - \\
 &  & pandas & 3.07 & 1.75 & - \\
 &  & Intercept-pandas Covariance & - & - & -0.631 \\
 & Database & Intercept & 0.734 & 0.856 & - \\
\midrule
\multirow{4}{*}{Qwen3-14B} & \multirow{3}{*}{Query} & Intercept & 6.58 & 2.57 & - \\
 &  & pandas & 4.33 & 2.08 & - \\
 &  & Intercept-pandas Covariance & - & - & -0.720 \\
 & Database & Intercept & 0.493 & 0.702 & - \\
\midrule
\multirow{4}{*}{Qwen3-8B} & \multirow{3}{*}{Query} & Intercept & 6.93 & 2.63 & - \\
 &  & pandas & 2.92 & 1.71 & - \\
 &  & Intercept-pandas Covariance & - & - & -0.684 \\
 & Database & Intercept & 0.555 & 0.745 & - \\
\midrule
\multirow{4}{*}{Qwen3-Coder-30B} & \multirow{3}{*}{Query} & Intercept & 8.58 & 2.93 & - \\
 &  & pandas & 5.55 & 2.35 & - \\
 &  & Intercept-pandas Covariance & - & - & -0.827 \\
 & Database & Intercept & 0.371 & 0.609 & - \\
\midrule
\multirow{4}{*}{GLM-4-32B} & \multirow{3}{*}{Query} & Intercept & 6.31 & 2.51 & - \\
 &  & pandas & 3.29 & 1.81 & - \\
 &  & Intercept-pandas Covariance & - & - & -0.512 \\
 & Database & Intercept & 1.10 & 1.05 & - \\
\bottomrule
\end{tabular}
    \vspace{-2mm}
    \caption{Random-effect summaries for the primary mixed-effects logistic regression, estimated separately for each LLM.}
\label{tab:model1-random-effects}
\end{table*}

\subsection{Detailed Logistic Regression Model Results: Structural Decomposition Model} \label{app:complexity_model_results}

Table~\ref{tab:model2-results-full} reports the full fixed-effect coefficient
table for the focused structural decomposition mixed-effects logistic regression, estimated separately for each of the nine LLMs.

\begin{table*}[!ht]
    \centering
    \scriptsize
    \setlength{\tabcolsep}{2.5pt}
    \resizebox{\textwidth}{!}{%
    \begin{tabular}{llrrrrrrrrrrrrrrrr}
    \toprule
        \textbf{Value} & \textbf{Model} & \textbf{(Intercept)} & \textbf{pandas} & \shortstack{\textbf{pandas $\times$}\\\textbf{Schema}} & \shortstack{\textbf{pandas $\times$}\\\textbf{SchemaLLM}} & \shortstack{\textbf{pandas $\times$}\\\textbf{PS}} & \textbf{Schema} & \textbf{SchemaLLM} & \textbf{PS} & \textbf{\texttt{SUBQUERY}} & \textbf{\texttt{COLUMN}} & \textbf{\texttt{SCALAR}} & \textbf{\texttt{JOIN}} & \shortstack{\textbf{pandas $\times$}\\\textbf{\texttt{SUBQUERY}}} & \shortstack{\textbf{pandas $\times$}\\\textbf{\texttt{COLUMN}}} & \shortstack{\textbf{pandas $\times$}\\\textbf{\texttt{SCALAR}}} & \shortstack{\textbf{pandas $\times$}\\\textbf{\texttt{JOIN}}} \\
    \midrule
        \multirow{9}{*}{Coefficient} & Gemini-2.5-Flash & 3.14 & -1.29 & -0.159 & -0.131 & -0.831 & 0.321 & 0.239 & 1.85 & -2.06 & -0.186 & -0.474 & -0.036 & 1.37 & -0.176 & 0.0986 & -0.00701 \\
         & Gemini-2.5-Pro & 4.01 & -1.63 & 0.297 & 0.365 & -0.651 & 0.05 & -0.0155 & 2.04 & -1.5 & -0.339 & -0.459 & 0.235 & 0.715 & -0.107 & 0.294 & -0.0713 \\
         & Gemma-4-26B & 1.61 & -0.0729 & 0.722 & 0.71 & 0.101 & -0.52 & -0.467 & 0.884 & -2.85 & 0.0213 & -0.0621 & -0.365 & 2.39 & -0.515 & -0.227 & 0.394 \\
         & GPT-5-Mini & 2.54 & -1.47 & -0.219 & -0.0785 & -0.537 & 0.0748 & -0.0636 & 2.23 & -1.03 & -0.0454 & 0.327 & -0.00685 & 0.181 & -0.234 & 0.221 & -0.0219 \\
         & GPT-5.1 & 2.73 & -1.73 & 0.0827 & 0.106 & -0.695 & 0.0604 & 0.00821 & 2.21 & -1.58 & -0.28 & 0.31 & 0.141 & 0.852 & 0.0758 & 0.264 & -0.287 \\
         & Qwen3-14B & 2.24 & -1.84 & -0.0791 & -0.109 & -0.899 & -0.0802 & -0.0348 & 1.74 & -1.48 & 0.0306 & -0.29 & -0.12 & 1.05 & -0.53 & -0.125 & -0.0552 \\
         & Qwen3-8B & 2.38 & -1.3 & -0.17 & -0.17 & -0.908 & -0.112 & -0.123 & 1.63 & -1.22 & -0.0358 & -0.462 & -0.124 & 0.816 & -0.313 & -0.002 & -0.0783 \\
         & Qwen3-Coder-30B & 2.74 & -2.49 & -0.000658 & 0.0253 & -0.973 & -0.228 & -0.173 & 1.79 & -1.57 & 0.069 & -0.643 & -0.17 & 0.83 & -0.516 & 0.349 & -0.0859 \\
         & GLM-4-32B & 2.1 & -1.05 & 0.193 & 0.355 & -0.609 & -0.456 & -0.531 & 1.69 & -1.74 & 0.289 & -0.673 & -0.476 & 1.15 & -0.494 & 0.54 & 0.119 \\
    \midrule
        \multirow{9}{*}{Standard Error} & Gemini-2.5-Flash & 0.369 & 0.123 & 0.095 & 0.0946 & 0.053 & 0.072 & 0.0717 & 0.0457 & 0.299 & 0.169 & 0.202 & 0.164 & 0.214 & 0.119 & 0.146 & 0.114 \\
         & Gemini-2.5-Pro & 0.39 & 0.134 & 0.104 & 0.104 & 0.0598 & 0.0782 & 0.078 & 0.0515 & 0.352 & 0.198 & 0.237 & 0.193 & 0.229 & 0.126 & 0.157 & 0.122 \\
         & Gemma-4-26B & 0.262 & 0.104 & 0.0822 & 0.0823 & 0.0395 & 0.0562 & 0.0563 & 0.0271 & 0.185 & 0.102 & 0.119 & 0.1 & 0.196 & 0.106 & 0.125 & 0.101 \\
         & GPT-5-Mini & 0.348 & 0.104 & 0.0956 & 0.0953 & 0.0535 & 0.0722 & 0.0719 & 0.0484 & 0.318 & 0.178 & 0.211 & 0.173 & 0.178 & 0.0988 & 0.12 & 0.0952 \\
         & GPT-5.1 & 0.328 & 0.116 & 0.096 & 0.0959 & 0.0557 & 0.0733 & 0.0732 & 0.0495 & 0.321 & 0.18 & 0.215 & 0.176 & 0.202 & 0.113 & 0.139 & 0.109 \\
         & Qwen3-14B & 0.251 & 0.116 & 0.0839 & 0.0839 & 0.0462 & 0.0645 & 0.0646 & 0.0399 & 0.243 & 0.136 & 0.161 & 0.132 & 0.215 & 0.12 & 0.141 & 0.115 \\
         & Qwen3-8B & 0.262 & 0.107 & 0.085 & 0.085 & 0.0444 & 0.065 & 0.065 & 0.0381 & 0.251 & 0.14 & 0.165 & 0.136 & 0.192 & 0.106 & 0.126 & 0.101 \\
         & Qwen3-Coder-30B & 0.241 & 0.128 & 0.0854 & 0.0855 & 0.0473 & 0.0673 & 0.0674 & 0.0415 & 0.275 & 0.153 & 0.181 & 0.148 & 0.24 & 0.133 & 0.157 & 0.127 \\
         & GLM-4-32B & 0.354 & 0.109 & 0.0852 & 0.0853 & 0.0443 & 0.0631 & 0.063 & 0.0359 & 0.239 & 0.133 & 0.157 & 0.13 & 0.198 & 0.108 & 0.129 & 0.103 \\
    \midrule
        \multirow{9}{*}{$p$-value} & Gemini-2.5-Flash & 1.83e-17 & 1.11e-25 & 0.0937 & 0.166 & 2.95e-55 & 8.19e-6 & 0.000844 & 0 & 6.47e-12 & 0.272 & 0.019 & 0.827 & 1.78e-10 & 0.138 & 0.499 & 0.951 \\
         & Gemini-2.5-Pro & 9.34e-25 & 2.84e-34 & 0.00418 & 0.000424 & 1.49e-27 & 0.523 & 0.842 & 0 & 2.08e-5 & 0.0866 & 0.0529 & 0.223 & 0.00185 & 0.398 & 0.061 & 0.559 \\
         & Gemma-4-26B & 8.92e-10 & 0.485 & 1.7e-18 & 6.25e-18 & 0.0109 & 2.18e-20 & 1.1e-16 & 2.67e-233 & 1.48e-53 & 0.835 & 0.602 & 0.000259 & 2.87e-34 & 1.05e-6 & 0.069 & 9.02e-5 \\
         & GPT-5-Mini & 3.27e-13 & 1.28e-45 & 0.022 & 0.41 & 9.18e-24 & 0.3 & 0.377 & 0 & 0.00116 & 0.798 & 0.122 & 0.969 & 0.308 & 0.018 & 0.0652 & 0.818 \\
         & GPT-5.1 & 9.23e-17 & 2.56e-50 & 0.389 & 0.268 & 1.23e-35 & 0.41 & 0.911 & 0 & 8.6e-7 & 0.12 & 0.15 & 0.423 & 2.49e-5 & 0.503 & 0.057 & 0.0083 \\
         & Qwen3-14B & 3.96e-19 & 1.49e-56 & 0.346 & 0.193 & 1.97e-84 & 0.214 & 0.59 & 0 & 1.08e-9 & 0.822 & 0.0714 & 0.365 & 9.46e-7 & 1.05e-5 & 0.375 & 0.63 \\
         & Qwen3-8B & 1.28e-19 & 2.36e-34 & 0.0457 & 0.0454 & 5.51e-93 & 0.0846 & 0.0584 & 0 & 1.08e-6 & 0.798 & 0.00525 & 0.361 & 2.03e-5 & 0.00297 & 0.987 & 0.437 \\
         & Qwen3-Coder-30B & 4.99e-30 & 4.7e-84 & 0.994 & 0.767 & 5.64e-94 & 0.000717 & 0.0105 & 0 & 9.91e-9 & 0.652 & 0.00038 & 0.25 & 0.00055 & 0.000108 & 0.0265 & 0.5 \\
         & GLM-4-32B & 3.11e-9 & 4.07e-22 & 0.0234 & 3.11e-5 & 4.97e-43 & 4.9e-13 & 3.89e-17 & 0 & 3.62e-13 & 0.0298 & 1.81e-5 & 0.000248 & 6.75e-9 & 4.31e-6 & 2.71e-5 & 0.249 \\
    \bottomrule
    \end{tabular}%
    }
    \vspace{-2mm}
    \caption{Full fixed-effect results for the focused structural decomposition mixed-effects logistic regression, estimated separately for each LLM. Logit-scale coefficients, standard errors, and raw $p$-values are reported in separate row groups. Subquery and scalar-function predictors are binary presence indicators; join and selected-column predictors are standardized log-transformed counts.}
    \label{tab:model2-results-full}
\end{table*}

Table~\ref{tab:model2-fit} reports the model fit summaries for the focused
structural decomposition mixed-effects logistic regression, including AIC, BIC,
log likelihood, and AUROC for each LLM-specific fit.

\begin{table*}[t]
\centering
\scriptsize
\setlength{\tabcolsep}{4pt}
\begin{tabular}{lrrrr}
\toprule
\textbf{Model} & \textbf{AIC} & \textbf{BIC} & \textbf{Log Likelihood} & \textbf{AUROC} \\
\midrule
Gemini-2.5-Flash & 20964.57 & 21128.86 & -10462.29 & 0.947 \\
Gemini-2.5-Pro & 18364.25 & 18528.53 & -9162.13 & 0.960 \\
Gemma-4-26B & 25506.36 & 25670.64 & -12733.18 & 0.917 \\
GPT-5-Mini & 20897.39 & 21061.68 & -10428.70 & 0.951 \\
GPT-5.1 & 20969.93 & 21134.21 & -10464.96 & 0.951 \\
Qwen3-14B & 25547.91 & 25712.19 & -12753.95 & 0.927 \\
Qwen3-8B & 24810.66 & 24974.94 & -12385.33 & 0.926 \\
Qwen3-Coder-30B & 25436.17 & 25600.46 & -12698.09 & 0.928 \\
GLM-4-32B & 24739.92 & 24904.21 & -12349.96 & 0.933 \\
\bottomrule
\end{tabular}%
    \vspace{-2mm}
    \caption{Model fit summaries for the focused structural decomposition mixed-effects logistic regression, estimated separately for each LLM.}
\label{tab:model2-fit}
\end{table*}

Table~\ref{tab:model2-random-effects} reports the random-effect summaries for
the focused structural decomposition mixed-effects logistic regression,
including query-level and database-level variance components for each
LLM-specific fit.

\begin{table*}[t]
\centering
\scriptsize
\setlength{\tabcolsep}{4pt}
\begin{tabular}{lllccc}
\toprule
\textbf{Model} & \textbf{Group} & \textbf{Component} & \textbf{Variance} & \textbf{SD} & \textbf{Correlation} \\
\midrule
\multirow{4}{*}{Gemini-2.5-Flash} & \multirow{3}{*}{Query} & Intercept & 9.96 & 3.16 & - \\
 &  & pandas & 3.48 & 1.86 & - \\
 &  & Intercept-pandas Covariance & - & - & -0.576 \\
 & Database & Intercept & 1.23 & 1.11 & - \\
\midrule
\multirow{4}{*}{Gemini-2.5-Pro} & \multirow{3}{*}{Query} & Intercept & 13.49 & 3.67 & - \\
 &  & pandas & 3.53 & 1.88 & - \\
 &  & Intercept-pandas Covariance & - & - & -0.539 \\
 & Database & Intercept & 1.30 & 1.14 & - \\
\midrule
\multirow{4}{*}{Gemma-4-26B} & \multirow{3}{*}{Query} & Intercept & 3.28 & 1.81 & - \\
 &  & pandas & 2.85 & 1.69 & - \\
 &  & Intercept-pandas Covariance & - & - & -0.108 \\
 & Database & Intercept & 0.661 & 0.813 & - \\
\midrule
\multirow{4}{*}{GPT-5-Mini} & \multirow{3}{*}{Query} & Intercept & 11.41 & 3.38 & - \\
 &  & pandas & 1.76 & 1.33 & - \\
 &  & Intercept-pandas Covariance & - & - & -0.647 \\
 & Database & Intercept & 1.05 & 1.02 & - \\
\midrule
\multirow{4}{*}{GPT-5.1} & \multirow{3}{*}{Query} & Intercept & 11.66 & 3.41 & - \\
 &  & pandas & 2.91 & 1.71 & - \\
 &  & Intercept-pandas Covariance & - & - & -0.623 \\
 & Database & Intercept & 0.889 & 0.943 & - \\
\midrule
\multirow{4}{*}{Qwen3-14B} & \multirow{3}{*}{Query} & Intercept & 6.45 & 2.54 & - \\
 &  & pandas & 4.05 & 2.01 & - \\
 &  & Intercept-pandas Covariance & - & - & -0.707 \\
 & Database & Intercept & 0.519 & 0.721 & - \\
\midrule
\multirow{4}{*}{Qwen3-8B} & \multirow{3}{*}{Query} & Intercept & 6.86 & 2.62 & - \\
 &  & pandas & 2.77 & 1.66 & - \\
 &  & Intercept-pandas Covariance & - & - & -0.670 \\
 & Database & Intercept & 0.574 & 0.758 & - \\
\midrule
\multirow{4}{*}{Qwen3-Coder-30B} & \multirow{3}{*}{Query} & Intercept & 8.34 & 2.89 & - \\
 &  & pandas & 5.26 & 2.29 & - \\
 &  & Intercept-pandas Covariance & - & - & -0.820 \\
 & Database & Intercept & 0.418 & 0.647 & - \\
\midrule
\multirow{4}{*}{GLM-4-32B} & \multirow{3}{*}{Query} & Intercept & 6.19 & 2.49 & - \\
 &  & pandas & 3.07 & 1.75 & - \\
 &  & Intercept-pandas Covariance & - & - & -0.497 \\
 & Database & Intercept & 1.22 & 1.10 & - \\
\bottomrule
\end{tabular}
    \vspace{-2mm}
    \caption{Random-effect summaries for the focused structural decomposition mixed-effects logistic regression, estimated separately for each LLM.}
\label{tab:model2-random-effects}
\end{table*}

\subsection{Schema-Token-Count Control for Persona Specificity}
\label{app:schema-token-control}

Because the expert persona condition is constructed to name gold schema and
value tokens, the Persona Score (PS) could in principle reduce to a
schema-leakage effect: prompts that copy more gold identifiers are trivially
easier to answer. We test this directly by asking whether PS retains
explanatory signal beyond the raw number of gold identifiers named in the
question.

We first note that schema availability is already a covariate in the primary
model: the \textit{Schema} and \textit{SchemaLLM} terms encode inclusion of the
DDL/catalog schema block and, for \textit{SchemaLLM}, additional LLM-generated
column and table descriptions. The expert rewrites do not disclose identifiers
the model otherwise lacked; they specify which of the already-available
identifiers are relevant. These schema covariates show only small,
model-dependent effects (Table~\ref{tab:model1-results-main}), so simply adding
schema information does not reliably improve performance.

We next implement an additional covariate to control for schema leakage. We define
\emph{raw schema-token count} as the number of gold SQL tokens (table names,
column names, and literal values) that appear in the question.
We add this count, together with a paradigm (SQL/pandas) interaction, to the primary per-model GLMM (App.~\ref{app:primary-model-spec}). Although PS and the raw token count are correlated ($r = 0.74$), their variance inflation factor is low ($\mathrm{VIF} = 2.24$), indicating that the two predictors carry enough independent variation to be estimated jointly.

Table~\ref{tab:schema-token-control} reports the results. After adding the
control, the PS coefficient remains positive and significant in 9/9 models
after FDR correction (all $p < 10^{-23}$). A likelihood-ratio test on the two PS
terms (main effect and paradigm interaction) confirms that they improve fit
beyond the raw token count in 9/9 models after FDR correction (all
$p < 10^{-47}$). The paradigm effect is likewise unaffected: the pandas
coefficient remains negative and significant in 8/9 models, with Gemma-4-26B the
only exception, matching the primary results.
These results indicate that Persona phrasing predicts accuracy beyond the number of gold identifiers named, which is contra-indicative of a pure schema-leakage explanation.

\begin{table}[t]
\centering
\small
\setlength{\tabcolsep}{4pt}
\begin{tabular}{lcccc}
\toprule
\textbf{Model} & \textbf{PS coef.} & \textbf{PS $p_{\mathrm{FDR}}$} & \textbf{LRT $\chi^2_2$} & \textbf{LRT $p_{\mathrm{FDR}}$} \\
\midrule
Gemini-2.5-Flash & 1.657 & $9{\times}10^{-59}$ & 389.5 & $6{\times}10^{-85}$ \\
Gemini-2.5-Pro   & 1.557 & $3{\times}10^{-43}$ & 283.4 & $4{\times}10^{-62}$ \\
Gemma-4-26B      & 0.682 & $1{\times}10^{-24}$ & 240.4 & $7{\times}10^{-53}$ \\
GPT-5-Mini       & 1.564 & $5{\times}10^{-50}$ & 456.7 & $6{\times}10^{-99}$ \\
GPT-5.1          & 1.361 & $2{\times}10^{-36}$ & 432.1 & $7{\times}10^{-94}$ \\
Qwen3-14B        & 1.155 & $4{\times}10^{-37}$ & 335.6 & $2{\times}10^{-73}$ \\
Qwen3-8B         & 1.014 & $2{\times}10^{-30}$ & 220.0 & $2{\times}10^{-48}$ \\
Qwen3-Coder-30B  & 1.112 & $3{\times}10^{-30}$ & 361.0 & $7{\times}10^{-79}$ \\
GLM-4-32B        & 1.337 & $6{\times}10^{-58}$ & 406.2 & $2{\times}10^{-88}$ \\
\bottomrule
\end{tabular}%
\vspace{-2mm}
\caption{Persona Score (PS) effect after controlling for the raw
schema-token count. Each row is a per-model primary GLMM
(App.~\ref{app:primary-model-spec}) augmented with a
\texttt{paradigm}\,$\times$\,raw-token-count term. \textbf{PS coef.} is the
PS main-effect log-odds coefficient; \textbf{PS $p_{\mathrm{FDR}}$} is its
Benjamini--Hochberg-corrected Wald $p$-value. \textbf{LRT} is a likelihood-ratio
test of the two PS terms (main effect and paradigm interaction) against the
model with only the raw-token-count control. PS remains positive and significant
in 9/9 models, and the LRT is significant in 9/9 models, indicating that persona
phrasing specificity predicts accuracy beyond the number of gold identifiers
named in the question.}
\vspace{-2mm}
\label{tab:schema-token-control}
\end{table}

\subsection{Persona Stratification Examples} \label{app:persona_examples}
Table~\ref{tab:persona-stratification-examples} shows representative
persona-stratified questions from the released artifact. The novice rewrites
preserve the user request in business-style language, while the expert rewrites
make the schema linkage, filters, joins, and output constraints explicit. The
PS values show the intended increase in schema specificity from novice to
expert phrasing.

\begin{table*}[t]
\centering
\scriptsize
\setlength{\tabcolsep}{4pt}
\renewcommand{\arraystretch}{1.08}
\begin{tabularx}{\textwidth}{@{}ll>{\raggedright\arraybackslash\ttfamily}X>{\raggedright\arraybackslash}X>{\raggedright\arraybackslash}X>{\raggedright\arraybackslash}Xc@{}}
\toprule
\textbf{ID} & \textbf{DB} & \textbf{Gold SQL} & \textbf{Original} & \textbf{Novice} & \textbf{Expert} & \textbf{PS: O/N/E} \\
\midrule
\texttt{916} &
\texttt{formula\_1} &
SELECT surname FROM drivers WHERE nationality = 'Italian' &
Please list the surnames of all the Italian drivers. &
What are the last names of the drivers from Italy? &
From the \texttt{drivers} table, list the \texttt{surname} values for every row where \texttt{nationality} equals \texttt{Italian}. &
0.833 / 0.670 / 1.000 \\
\addlinespace
\texttt{492} &
\texttt{card\_games} &
SELECT type FROM sets WHERE name LIKE '\%FROM the Vault: Lore\%' &
For the set ``From the Vault: Lore'', what is its expansion type? &
What kind of expansion is ``From the Vault: Lore''? &
In the \texttt{sets} table, find the row where \texttt{name} contains \texttt{FROM the Vault: Lore}, and return only the \texttt{type} column as the expansion type. &
0.730 / 0.403 / 0.944 \\
\addlinespace
\texttt{1377} &
\texttt{student\_club} &
SELECT COUNT(member\_id) FROM member WHERE position = 'President' &
How many student have the position of president? &
How many students are presidents? &
From the \texttt{member} table, count rows where \texttt{position} equals \texttt{President}, using non-null \texttt{member\_id} values. &
0.619 / 0.425 / 0.972 \\
\bottomrule
\end{tabularx}
\caption{Examples of original, novice, and expert persona phrasings with the
corresponding gold SQL. O/N/E denotes the Persona Score for the original,
novice, and expert question, respectively.}
\label{tab:persona-stratification-examples}
\end{table*}

\end{document}